%% file: BBOB2012_saACMES_noisy.tex
\documentclass{sig-alternate}
\usepackage{graphicx}
\usepackage{tabularx}
\usepackage{xcolor}
\usepackage{float}
\usepackage{rotating}
\usepackage{xstring} 

\newcommand{\bbobdatapath}{ppdata/}

\input{\bbobdatapath bbob_pproc_commands.tex}

\graphicspath{{\bbobdatapath}}
 \newcommand{\algaperfprof}{IPOP-aCMA}
 \newcommand{\algbperfprof}{IPOP-$^{s\ast}$aACM}
% ...

\newcommand{\ERT}{\ensuremath{\mathrm{ERT}}}

\newcommand{\Df}{\ensuremath{\Delta f}}

\newcommand{\fopt}{\ensuremath{f_\mathrm{opt}}}
\newcommand{\ftarget}{\ensuremath{f_\mathrm{t}}}

\begin{document}

%
\conferenceinfo{GECCO'12,} {July 7--11, 2012, Philadelphia, USA.}
\CopyrightYear{2012}
\crdata{978-1-4503-0073-5/10/07}
\clubpenalty=10000
\widowpenalty = 10000

\title{Black-box optimization benchmarking of IPOP-saACM-ES on the BBOB-2012 noisy testbed}

%
%
%
%
%

\numberofauthors{3}
 \author{
 \alignauthor
 Ilya Loshchilov \\
 \affaddr{TAO, INRIA Saclay}\\
 \affaddr{U. Paris Sud, F-91405 Orsay}\\
 \and 
 \alignauthor
  Marc Schoenauer\\
 \affaddr{TAO, INRIA Saclay}\\
 \affaddr{U. Paris Sud, F-91405 Orsay}\\
 \email{firstname.lastname@inria.fr}
 \and 
 \alignauthor
  Mich\`ele Sebag\\
 \affaddr{CNRS, LRI UMR 8623}\\
 \affaddr{U. Paris Sud, F-91405 Orsay}\\
}

\def\SAM{$^{s\ast}$}
\def\IPOPsaACM{IPOP-\SAM\hspace{-0.50ex}aACM-ES}
\def\saACM{\SAM\hspace{-0.50ex}aACM-ES}
\def\sACM{\SAM\hspace{-0.50ex}ACM-ES}
\def\BIPOPsaACM{BIPOP-\SAM\hspace{-0.50ex}aACM-ES}
\def\IPOPCMA{IPOP-aCMA-ES}
\def\BIPOPCMA{BIPOP-CMA-ES}
\def\mulCMA{$(\mu / \mu_w , \lambda)$-CMA-ES}
\newcommand{\R}{\mathbb{R}}
\newcommand{\Rd}{\R^{D}}
\newcommand{\NormalNull}[1]{{\mathcal N}\hspace{-0.13em}\left(\ve{0},#1\right)}

\maketitle
\begin{abstract}
  In this paper, we study the performance of \IPOPsaACM, recently proposed self-adaptive surrogate-assisted Covariance Matrix Adaptation Evolution Strategy.
	The algorithm was tested using restarts till a total number of function evaluations of $10^6D$ was reached, where $D$ is the dimension of the function search space. 
	The experiments show that the surrogate model control allows \IPOPsaACM\ to be as robust as the original IPOP-aCMA-ES and 
	outperforms the latter by a factor from 2 to 3 on 6 benchmark problems with moderate noise.
	On 15 out of 30 benchmark problems in dimension 20, \IPOPsaACM\ exceeds the records observed during BBOB-2009 and BBOB-2010.
	
\end{abstract}




\section{Introduction}

When dealing with expensive optimization objectives, the surrogate-assisted approaches proceed by learning
a surrogate model of the objective, and using this surrogate to reduce the number of computations of the objective
function in various ways.

Many surrogate modelling approaches have been used within 
Evolution Strategies (ESs) and Covariance Matrix Adaptation Evolution Strategy (CMA-ES): 
Radial Basis Functions network \cite{Hoffmann2006IEEE}, Gaussian Processes \cite{Ulmer2003CEC}, Artificial
Neural Network \cite{YJin2005}, Support Vector Regression \cite{KramerInformatica2010}, 
Local-Weighted Regression \cite{kernHansenMetaPPSN06,augerEvoNum2010},
Ranking Support Vector Machine (Ranking SVM) \cite{runarssonPPSN06,rankSurrogatePPSN10,Runarsson2011ISDA}.
In most cases, the surrogate model is used as a filter (to select $\lambda_{Pre}$ promising pre-children) and/or to estimate the fitness of some individuals in the current population.
An example of surrogate-assisted CMA-ES with filtering strategy can be found in \cite{rankSurrogatePPSN10}.

A well-known drawback of surrogate-assisted optimization is a strong dependence of the results on hyper-parameters used to build the surrogate model. 
Some optimal settings of hyper-parameters for a specific set of problems can be found by offline tuning, however for a new problem they are unknown in the black-box scenario. Moreover, the optimal hyper-parameters may dynamically change during the optimization of the function. 

Motivated by this open issues, new self-adapted surrogate-assisted \saACM\ algorithm have been proposed combining surrogate-assisted optimization of the expensive function and online optimization of the surrogate model hyper-parameters \cite{ACMGECCO2012}.

\section{The Algorithms}

\subsection{The \mulCMA }

In each iteration $t$, \mulCMA\ \cite{HansenECJ01} samples $\lambda$ new solutions $x_i \in \Rd$, where $i=1,\ldots,\lambda$, and selects the best $\mu$ among them. 
These $\mu$ points update the distribution of parameters of the algorithm to increase the probability of successful steps in iteration $t+1$.
The sampling is defined by a multi-variate normal distribution, $\mathcal N(\textbf{$m$}^t,{\sigma^t}^2C^t)$, with current mean of distribution $\textbf{$m$}^t$, $D\times D$ covariance matrix $C^t$ and step-size ${\sigma^t}$.

The active version of the CMA-ES proposed in \cite{1830788,2006:JastrebskiArnold} introduces a weighted negative update of the covariance matrix taking into account the information about $\lambda-\mu$ worst points as well as about $\mu$ best ones. The new version improves CMA-ES on 9 out of 12 tested unimodal functions by a factor up to 2, and the advantages are more pronounced in larger dimension. While the new update scheme does not guarantee the positive-definiteness of the covariance matrix, it can be numerically controlled \cite{1830788}. 
Since in our study we do not observe any negative effects of this issue, we will use aCMA-ES, the active version of the CMA-ES, for comparison with the surrogate-assisted algorithms.

\subsection{The \sACM }

The \sACM\ \cite{ACMGECCO2012} is the surrogate-assisted version of the \mulCMA, where the surrogate model is used periodically instead of the expensive function for direct optimization.
The use of Ranking SVM allows to preserve the property of CMA-ES of invariance with respect to rank-preserving transformation of the fitness function. The property of invariance with respect to the orthogonal transformation of the search space is preserved thanks to the definition of the kernel function by the covariance matrix, adapted during the search. 

In \sACM\ we perform the following surrogate-assisted optimization loop: we optimize the surrogate model $\hat{f}$ for $\hat{n}$ generations by the CMA-ES, then we continue and optimize the expensive function $f(x)$ for one generation. To adjust the number of generations $\hat{n}$ for the next time, the model error can be computed as a fraction of incorrectly predicted comparison relations that we observe, when we compare the ranking of the last $\lambda$ evaluated points according to $f(x)$ and $\hat{f}$. The \sACM\ uses the generation of the CMA-ES as a black-box procedure, and it has been shown in \cite{ACMGECCO2012}, that the improvement of the CMA-ES from passive to active version (aCMA-ES) leads to a comparable improvement of its surrogate-assisted versions (\sACM\ and \saACM).

The main novelty of the \sACM\ is the online optimization of the surrogate model hyper-parameters during the optimization of the fitness function. The algorithm performs the search in the space of model hyper-parameters, generating $\lambda_{hyp}$ surrogate models in each iteration. The fitness of the model can be measured as a prediction error of the ranking on the last $\lambda$ evaluated points. This allows the user to define only the range of hyper-parameters and let algorithm to find the most suitable values for the current iteration $t$. 

The detailed description of \sACM\ is given in \cite{ACMGECCO2012}.

\subsection{The Benchmarked Algorithms }

For benchmarking we consider \sACM\ in IPOP restart scenario (\IPOPsaACM) using default parameters and termination criteria as given in \cite{1830788} and \cite{ACMGECCO2012}.
The only one parameter of the surrogate part of \IPOPsaACM\ different from the default one is 
the index of generation $g_{start}$ when we start to use the surrogate model. 
For noisy case we set $g_{start}=5 (i_{restart}+1)$ instead of default $g_{start}=10$.
We found that sometimes it makes sense to postpone the surrogate-assisted search if several restarts ($i_{restart}$) were performed.

%
\section{Results}

\begin{figure*}
\centering
\begin{tabular}{@{}c@{}c@{}c@{}c@{}c@{}}
\includegraphics[width=0.2\textwidth, trim=20mm 7mm 15mm 3mm, clip]{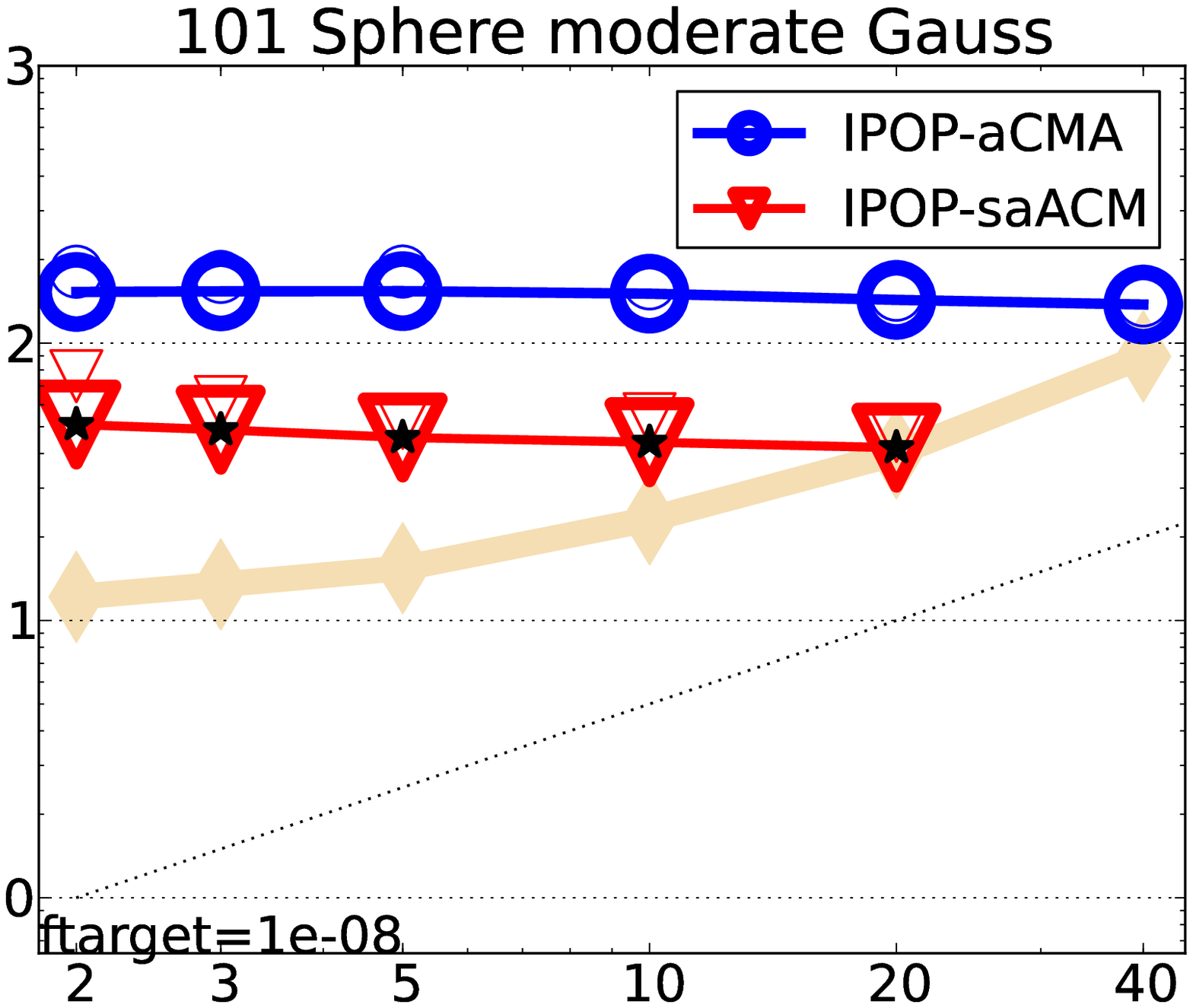}&
\includegraphics[width=0.2\textwidth, trim=20mm 7mm 15mm 3mm, clip]{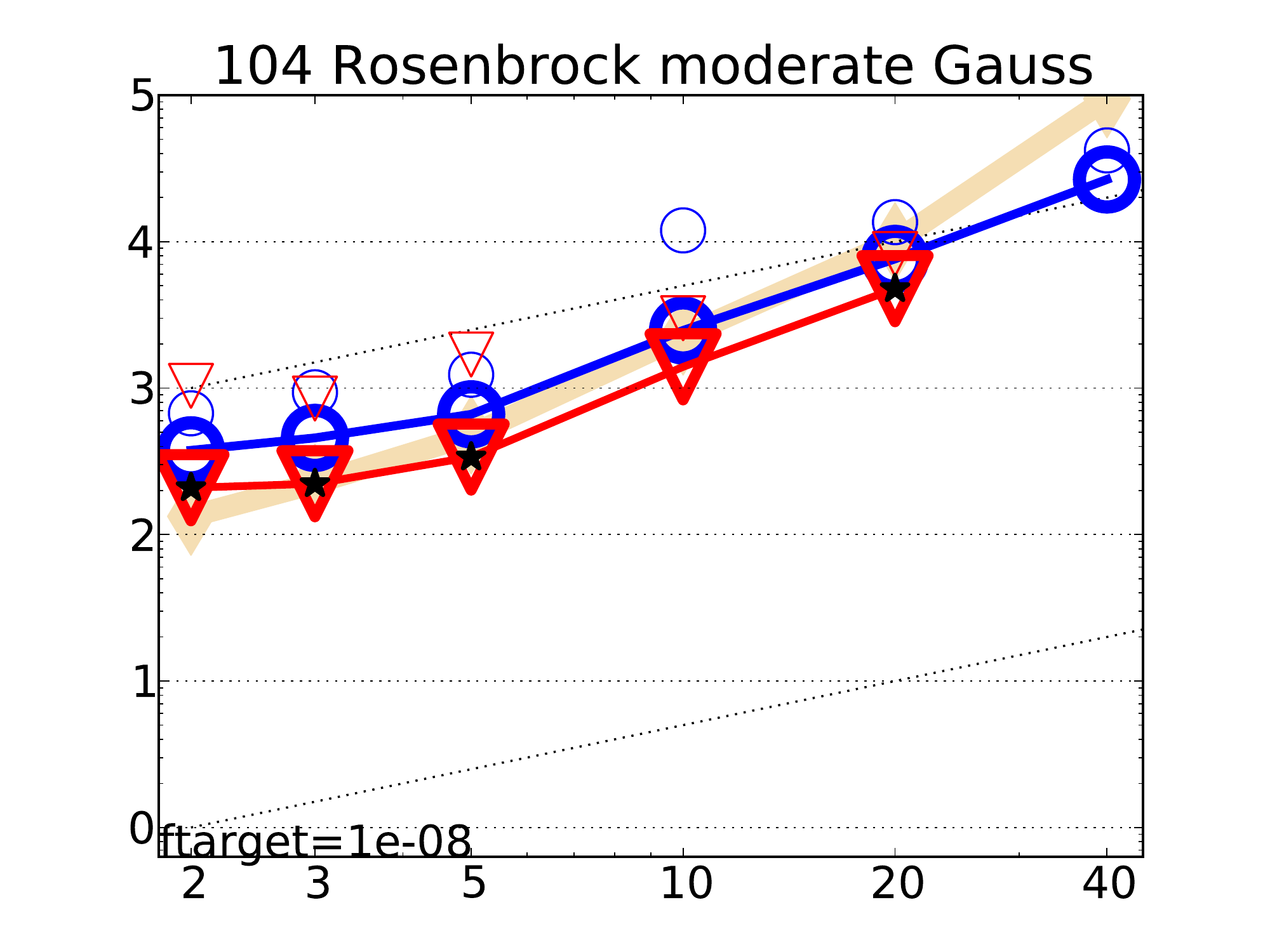}&
\includegraphics[width=0.2\textwidth, trim=20mm 7mm 15mm 3mm, clip]{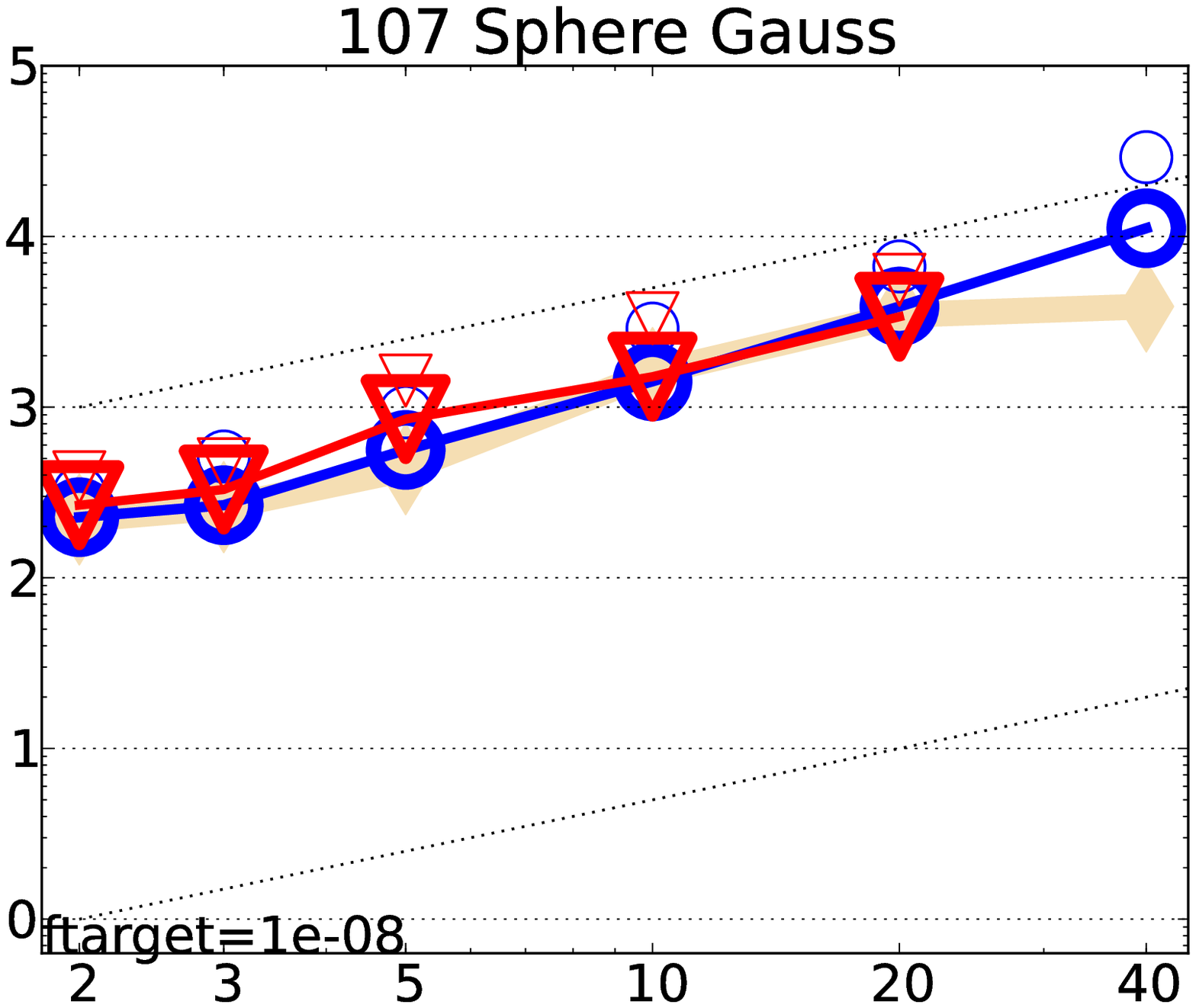}&
\includegraphics[width=0.2\textwidth, trim=20mm 7mm 15mm 3mm, clip]{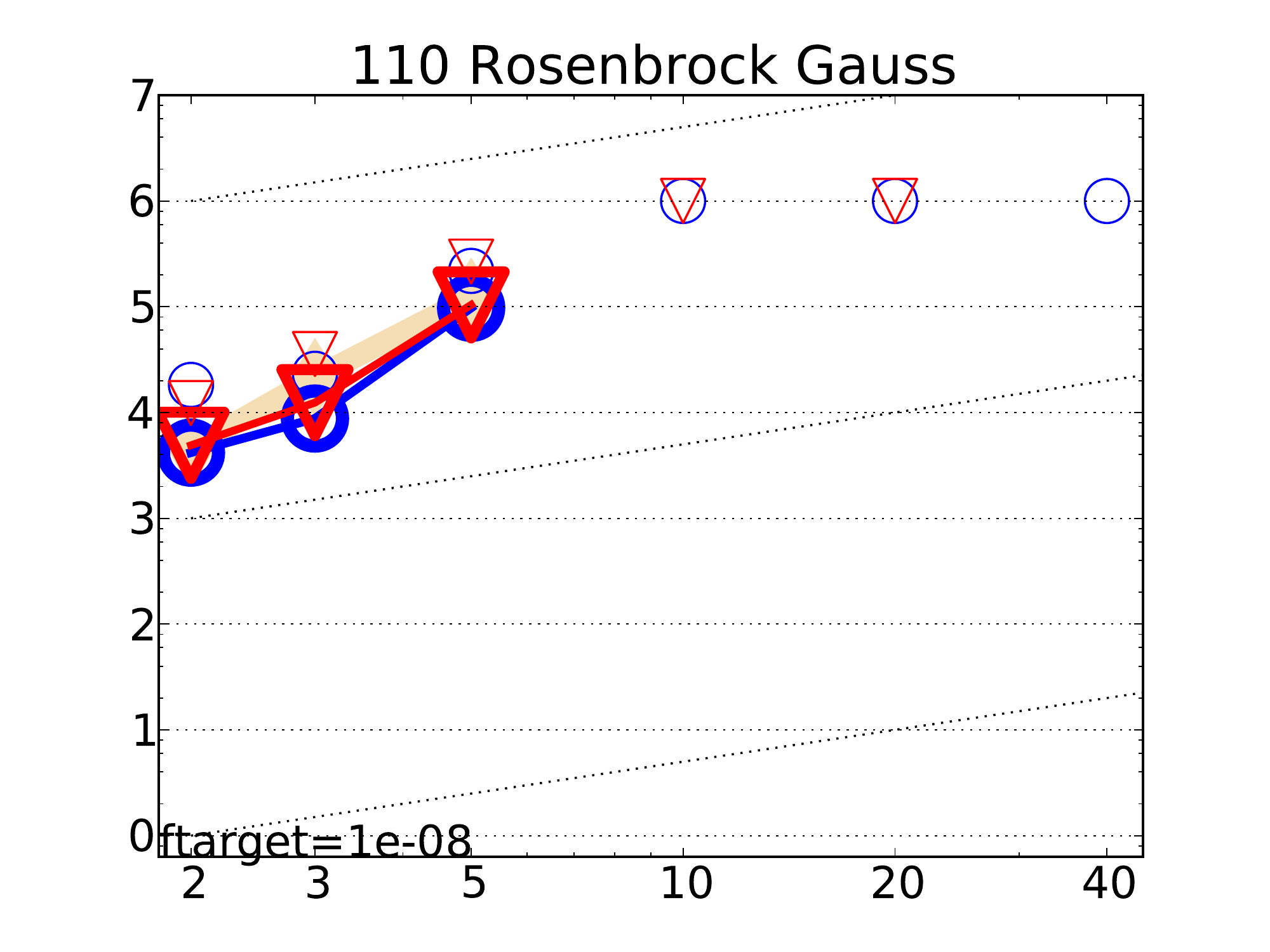}&
\includegraphics[width=0.2\textwidth, trim=20mm 7mm 15mm 3mm, clip]{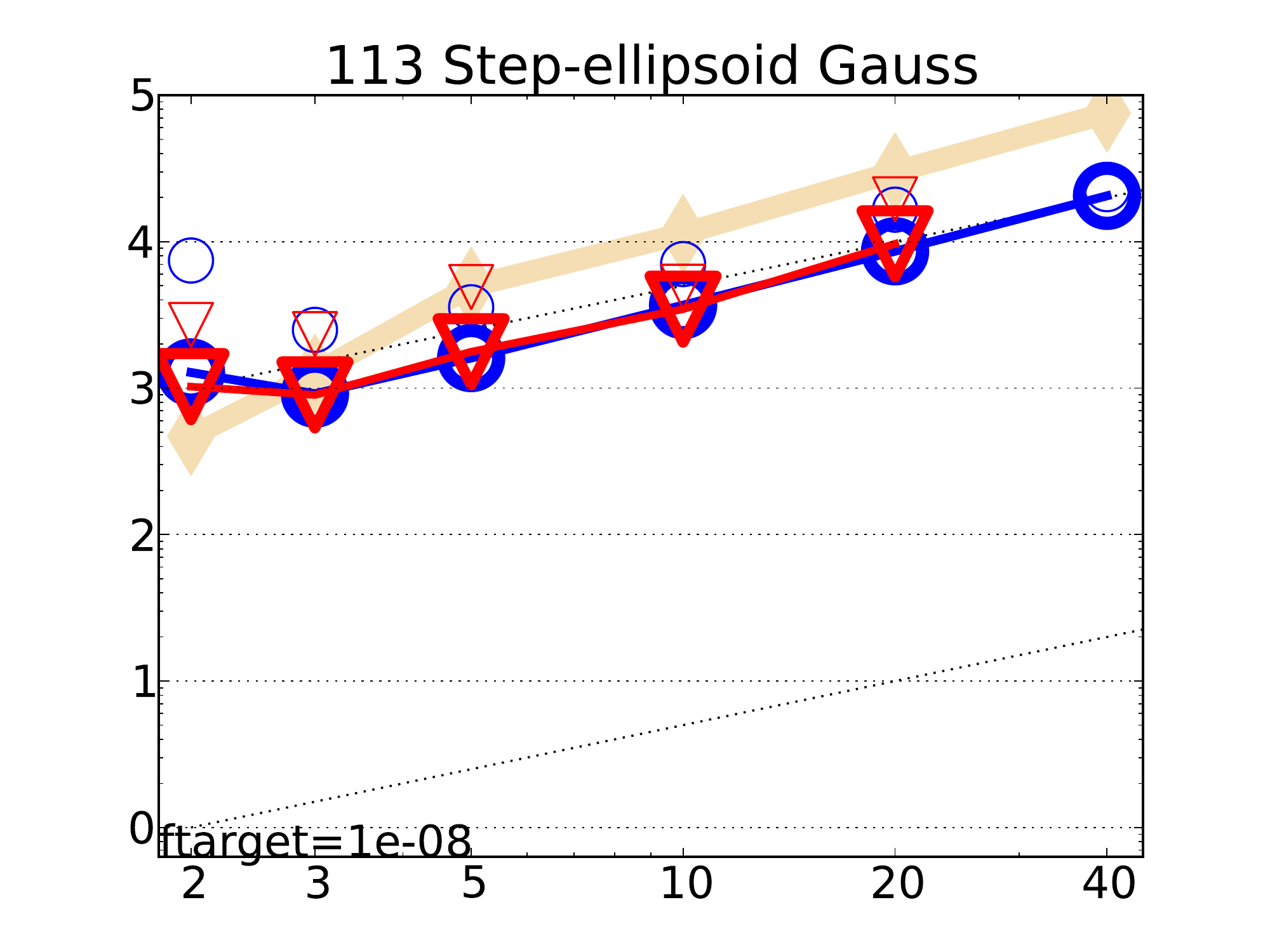}\\
\includegraphics[width=0.2\textwidth, trim=20mm 7mm 15mm 3mm, clip]{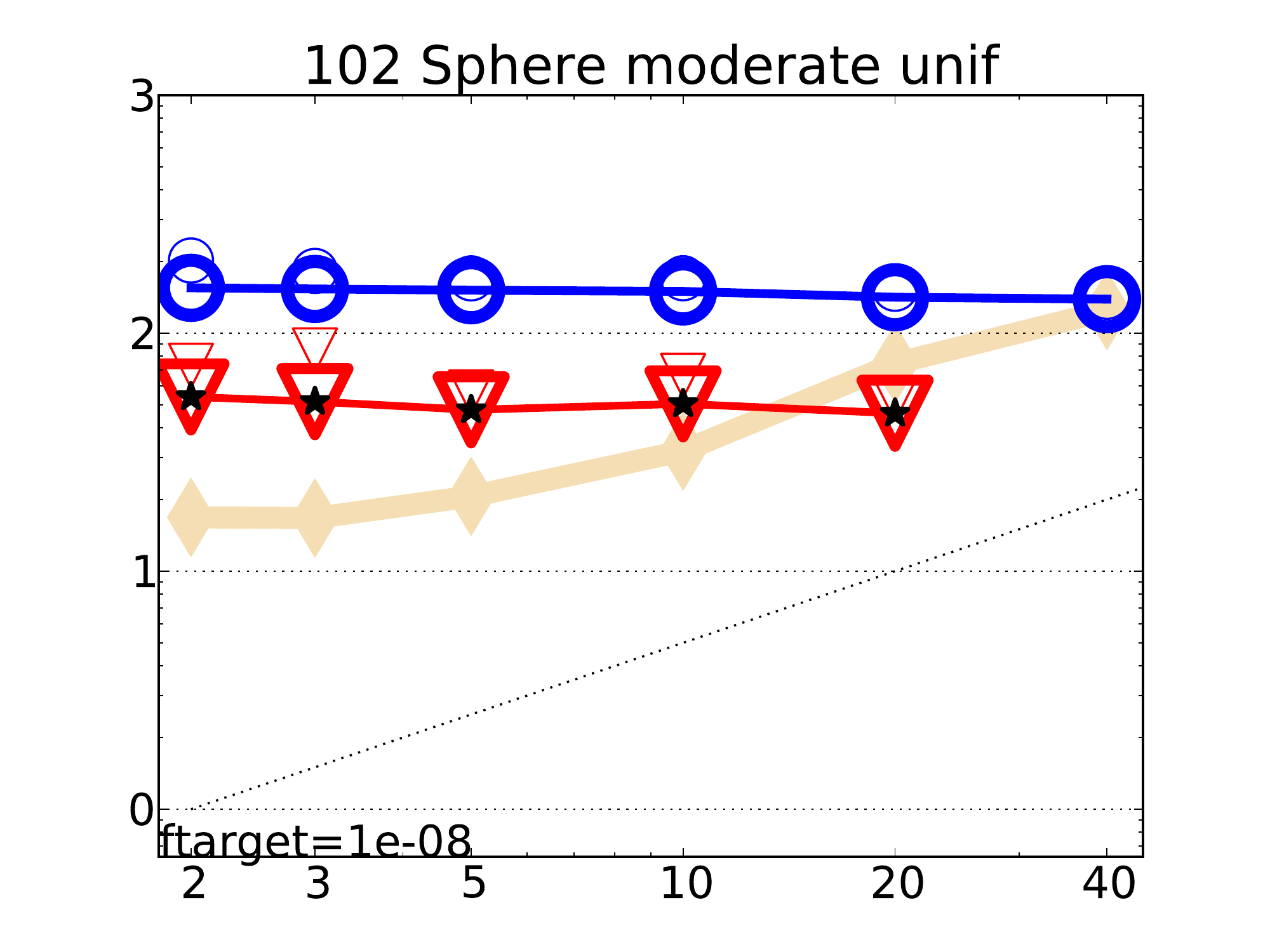}&
\includegraphics[width=0.2\textwidth, trim=20mm 7mm 15mm 3mm, clip]{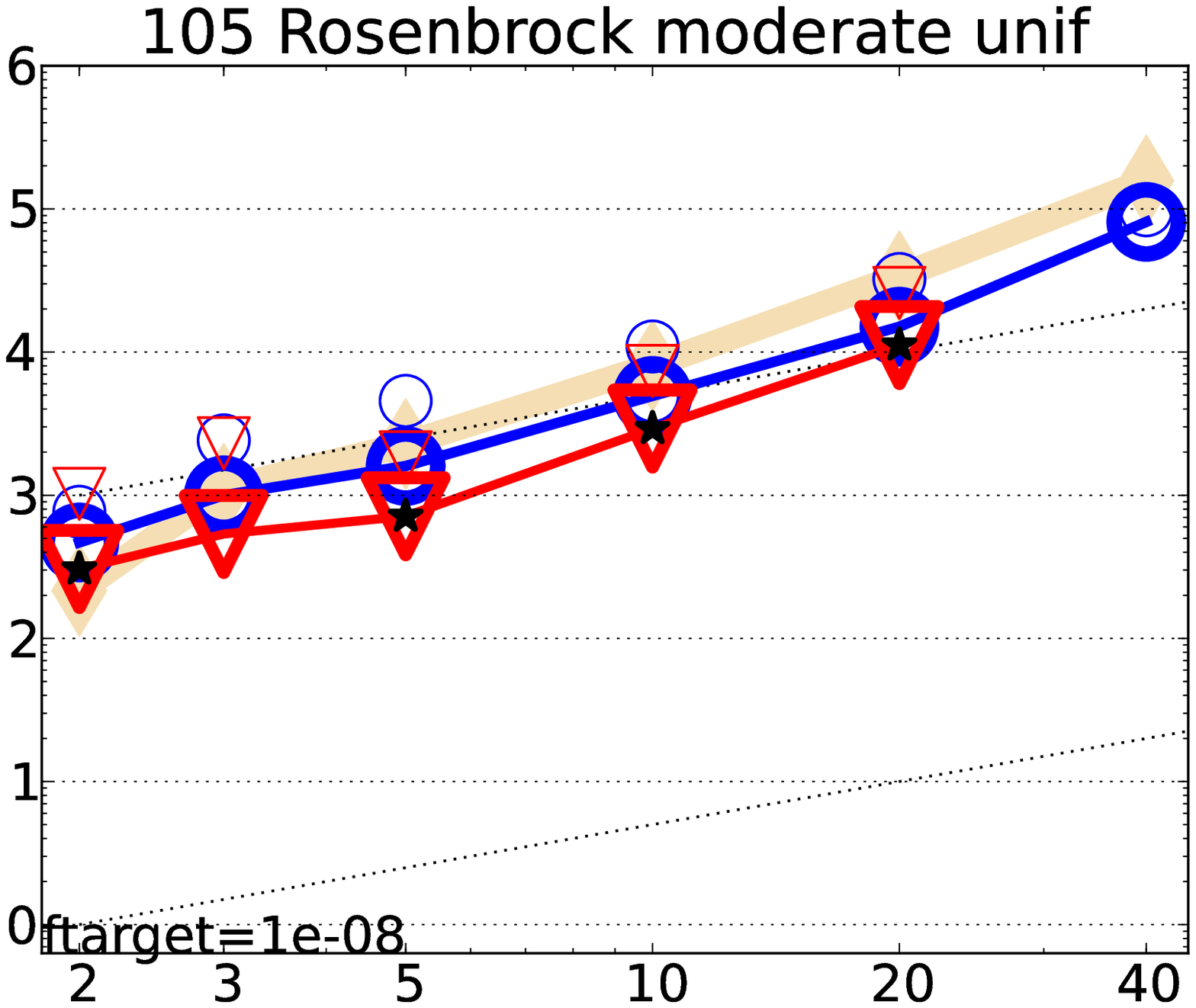}&
\includegraphics[width=0.2\textwidth, trim=20mm 7mm 15mm 3mm, clip]{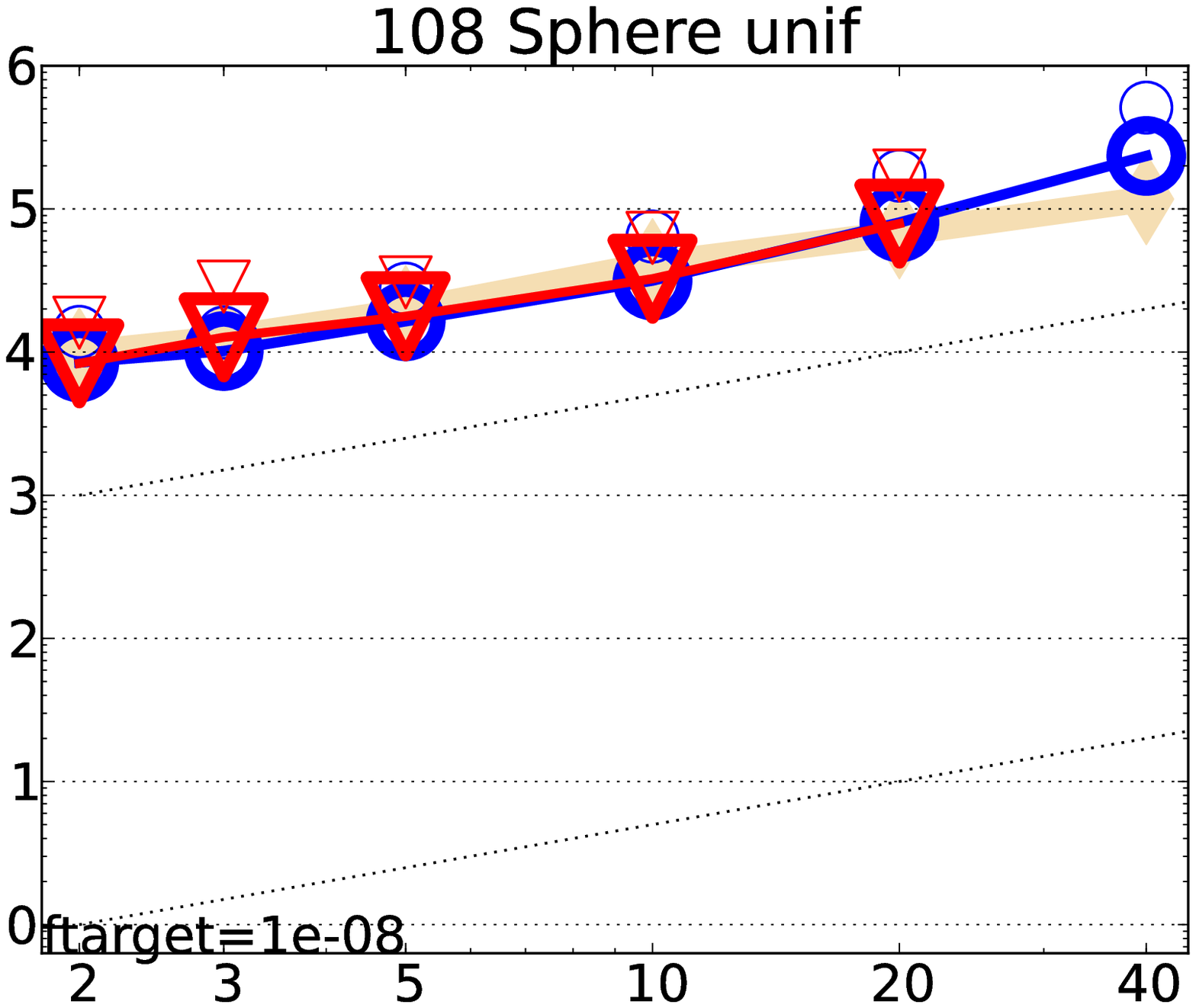}&
\includegraphics[width=0.2\textwidth, trim=20mm 7mm 15mm 3mm, clip]{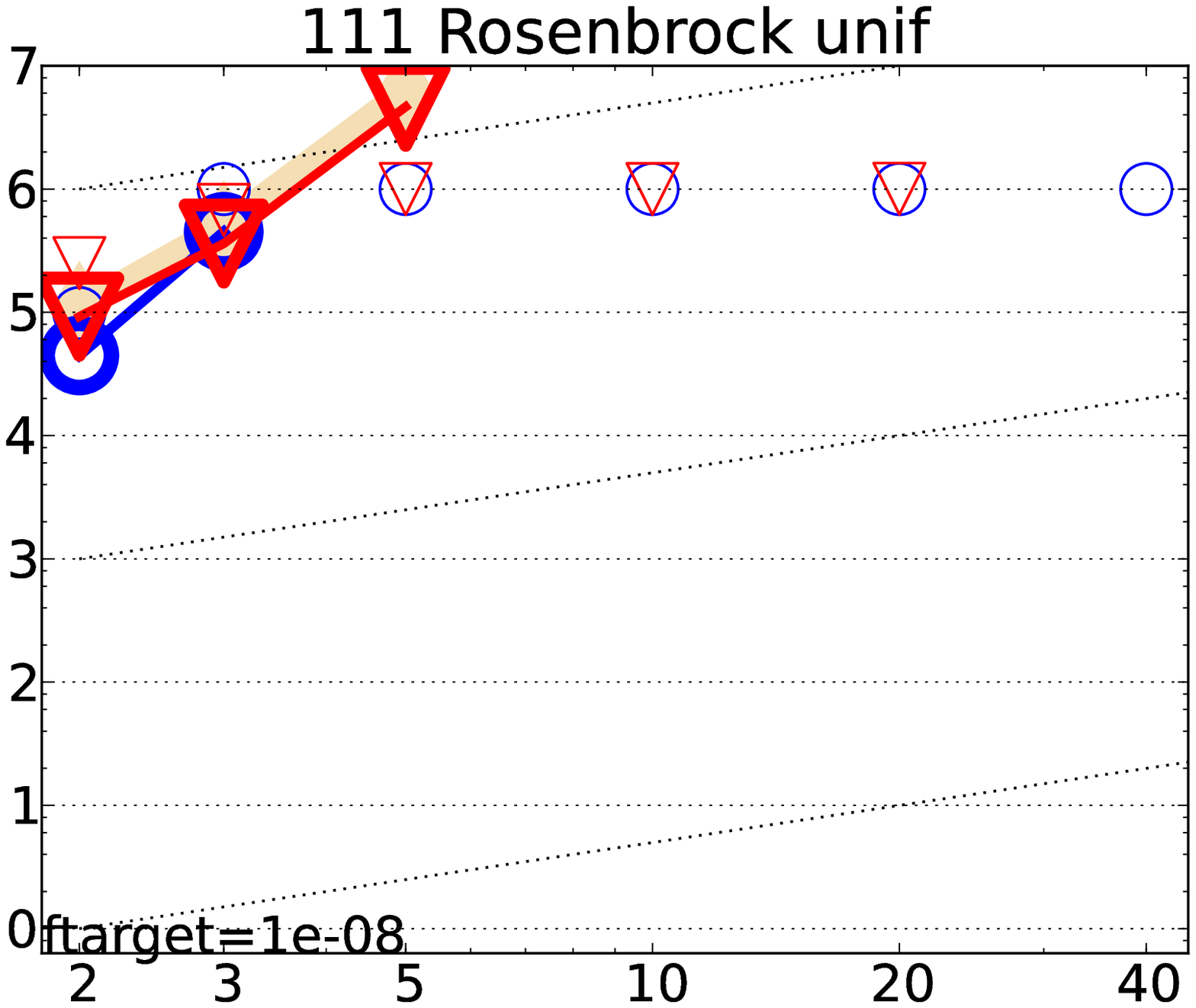}&
\includegraphics[width=0.2\textwidth, trim=20mm 7mm 15mm 3mm, clip]{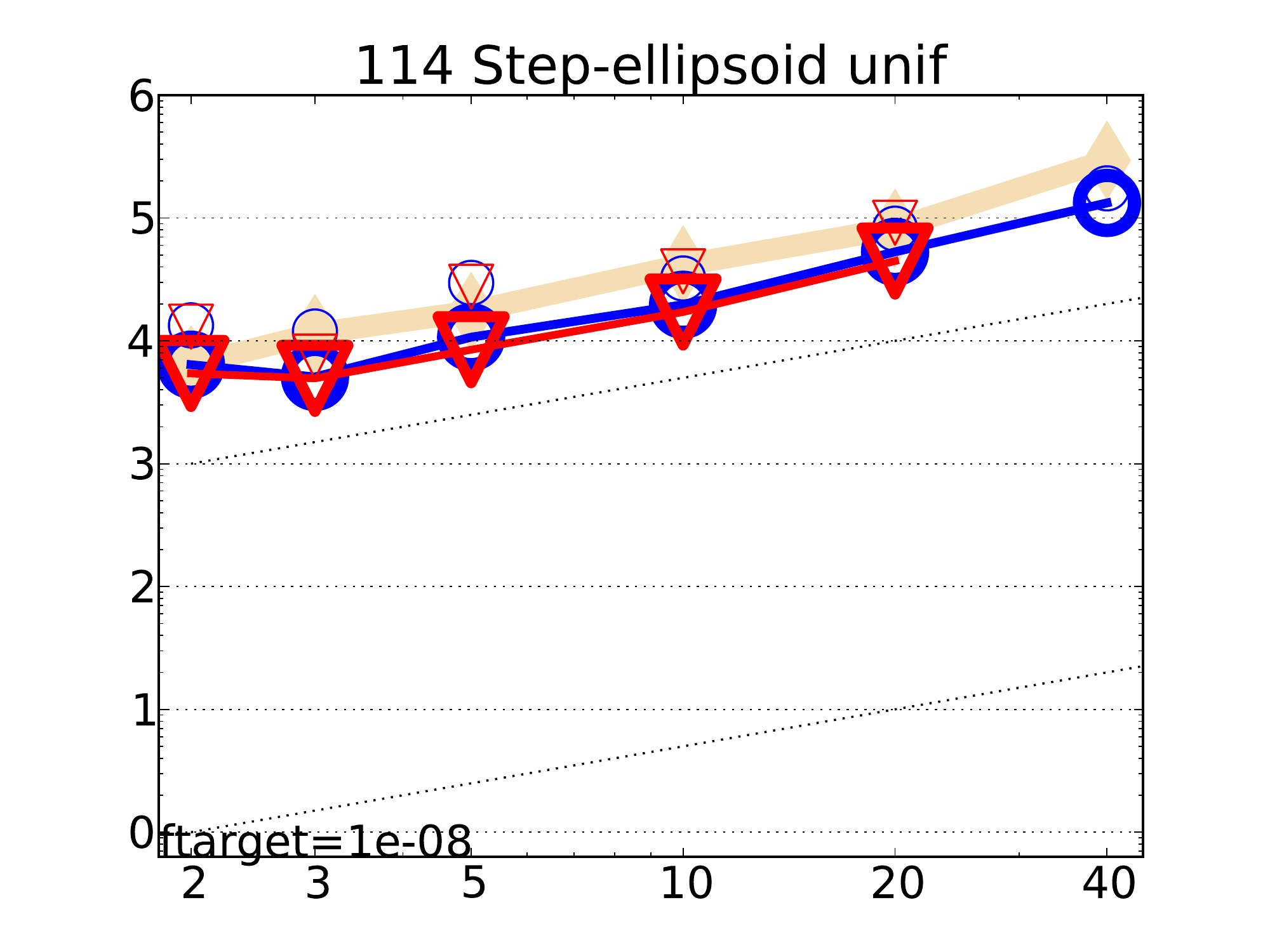}\\
\includegraphics[width=0.2\textwidth, trim=20mm 7mm 15mm 3mm, clip]{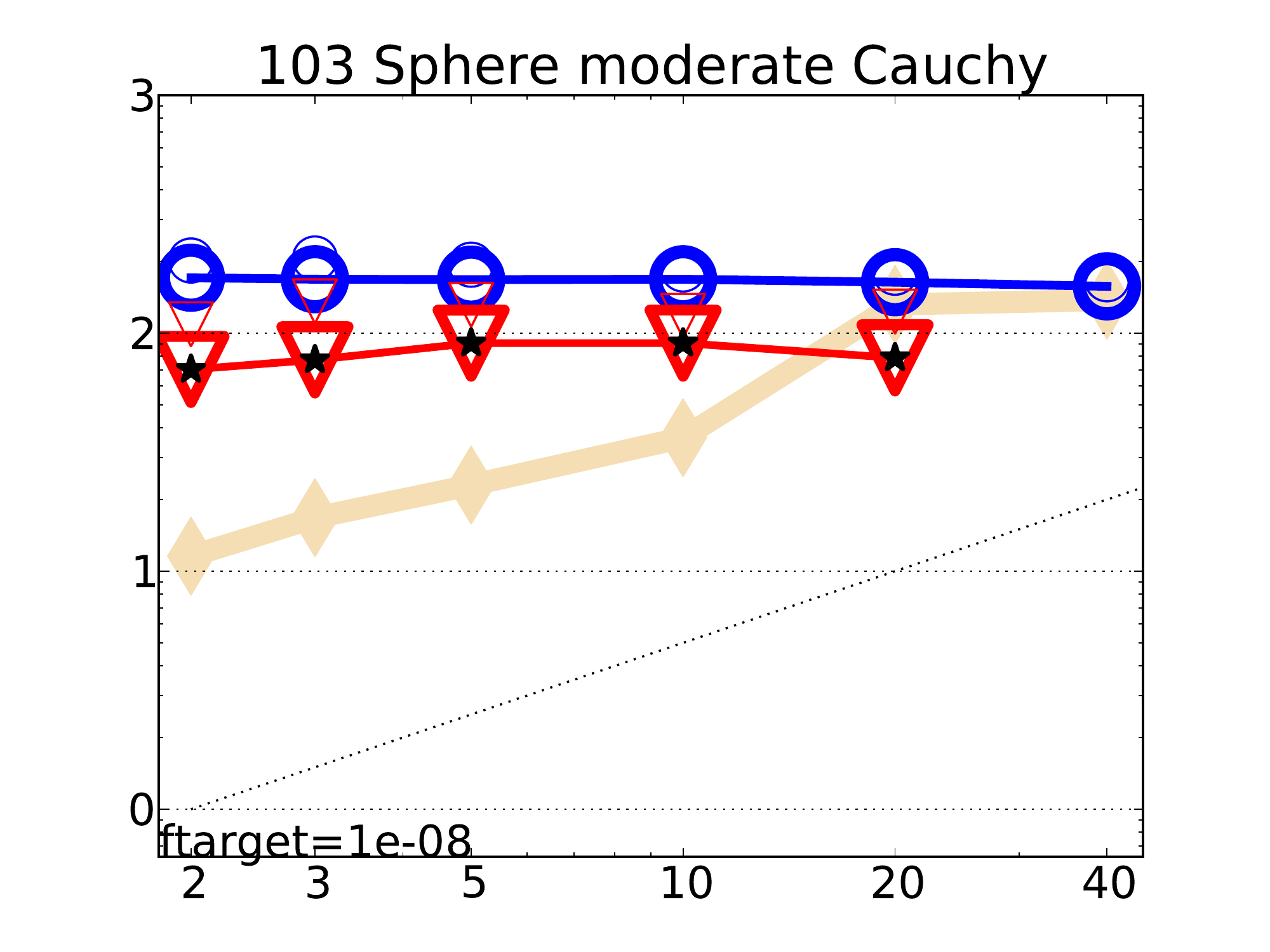}&
\includegraphics[width=0.2\textwidth, trim=20mm 7mm 15mm 3mm, clip]{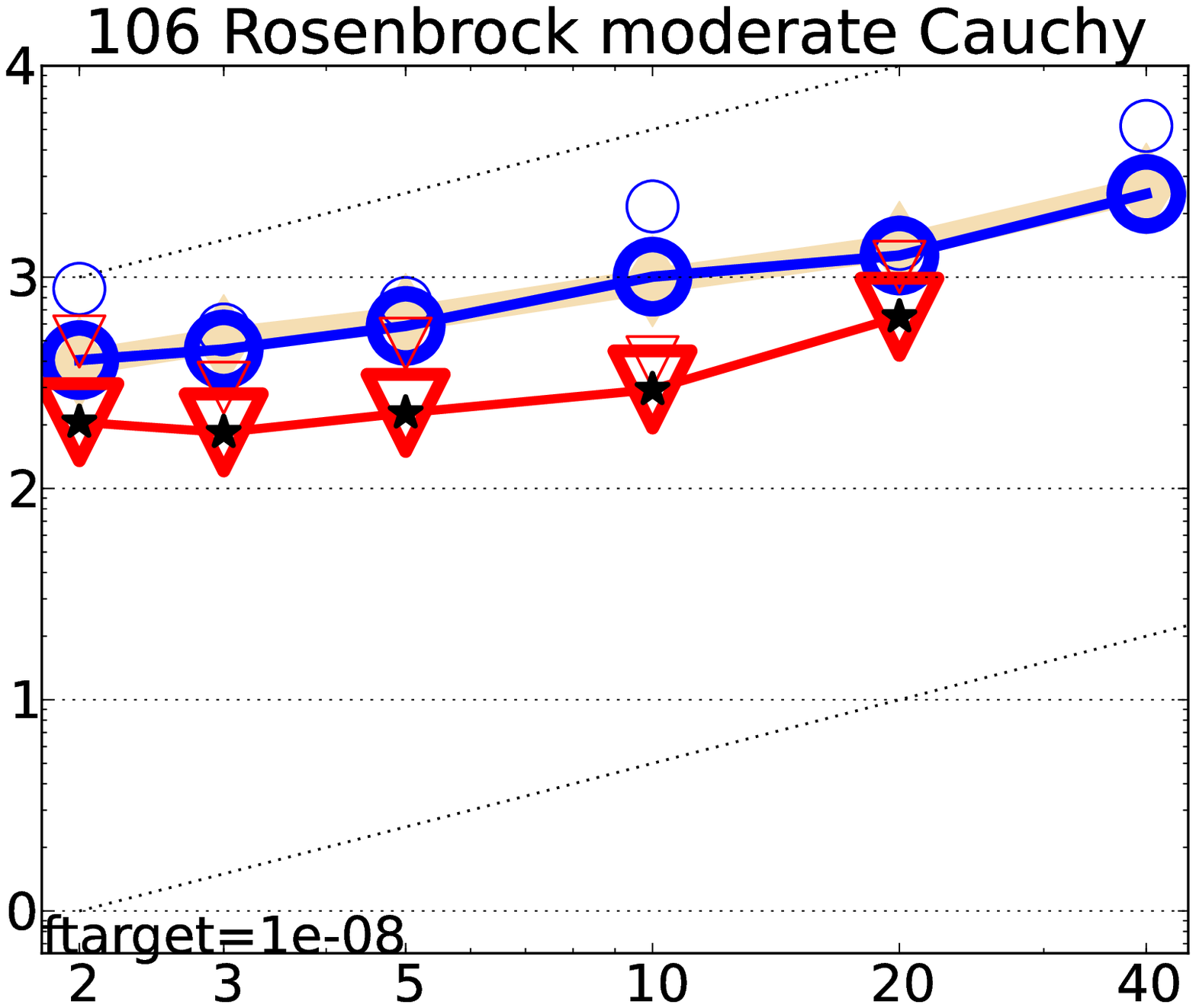}&
\includegraphics[width=0.2\textwidth, trim=20mm 7mm 15mm 3mm, clip]{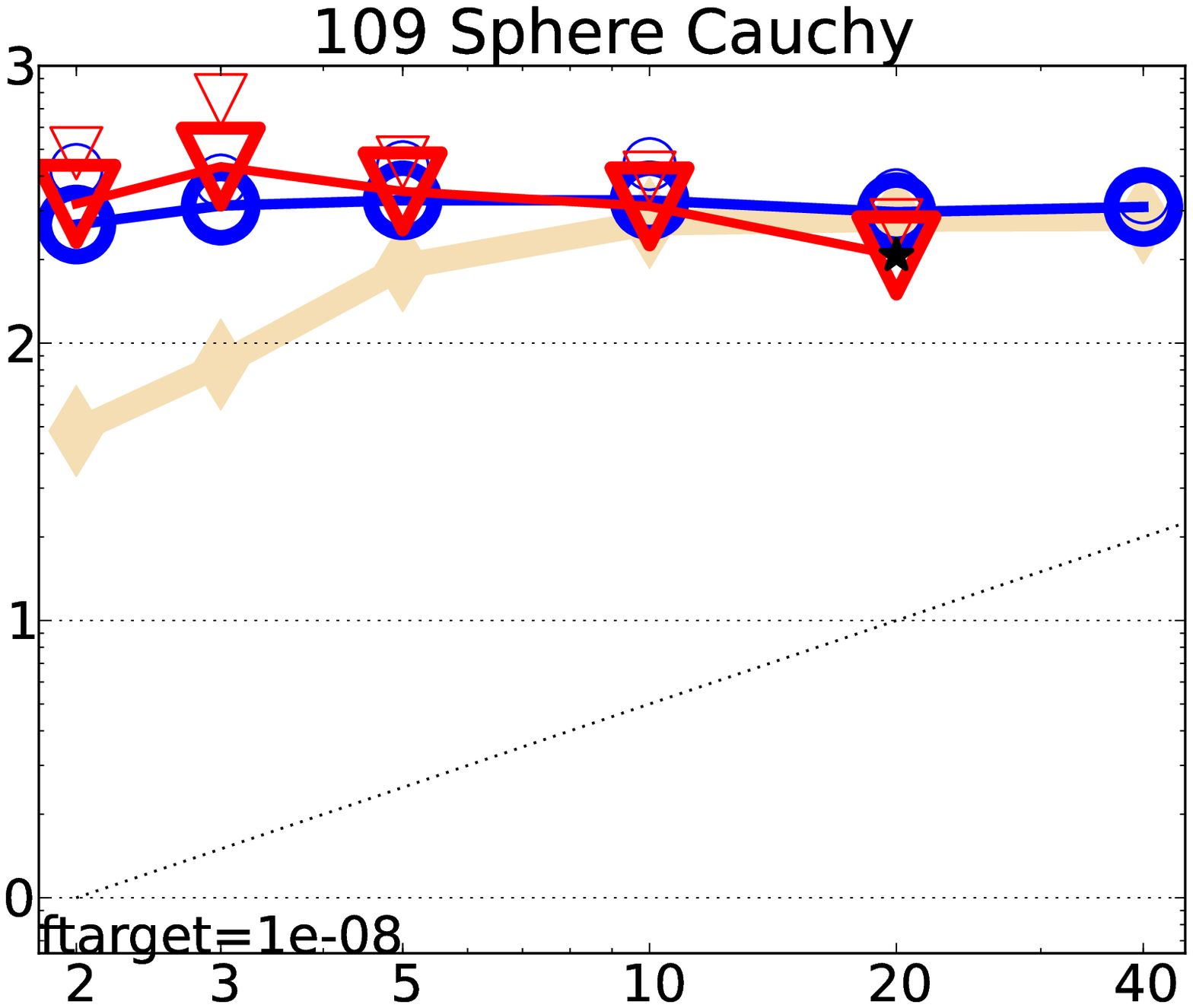}&
\includegraphics[width=0.2\textwidth, trim=20mm 7mm 15mm 3mm, clip]{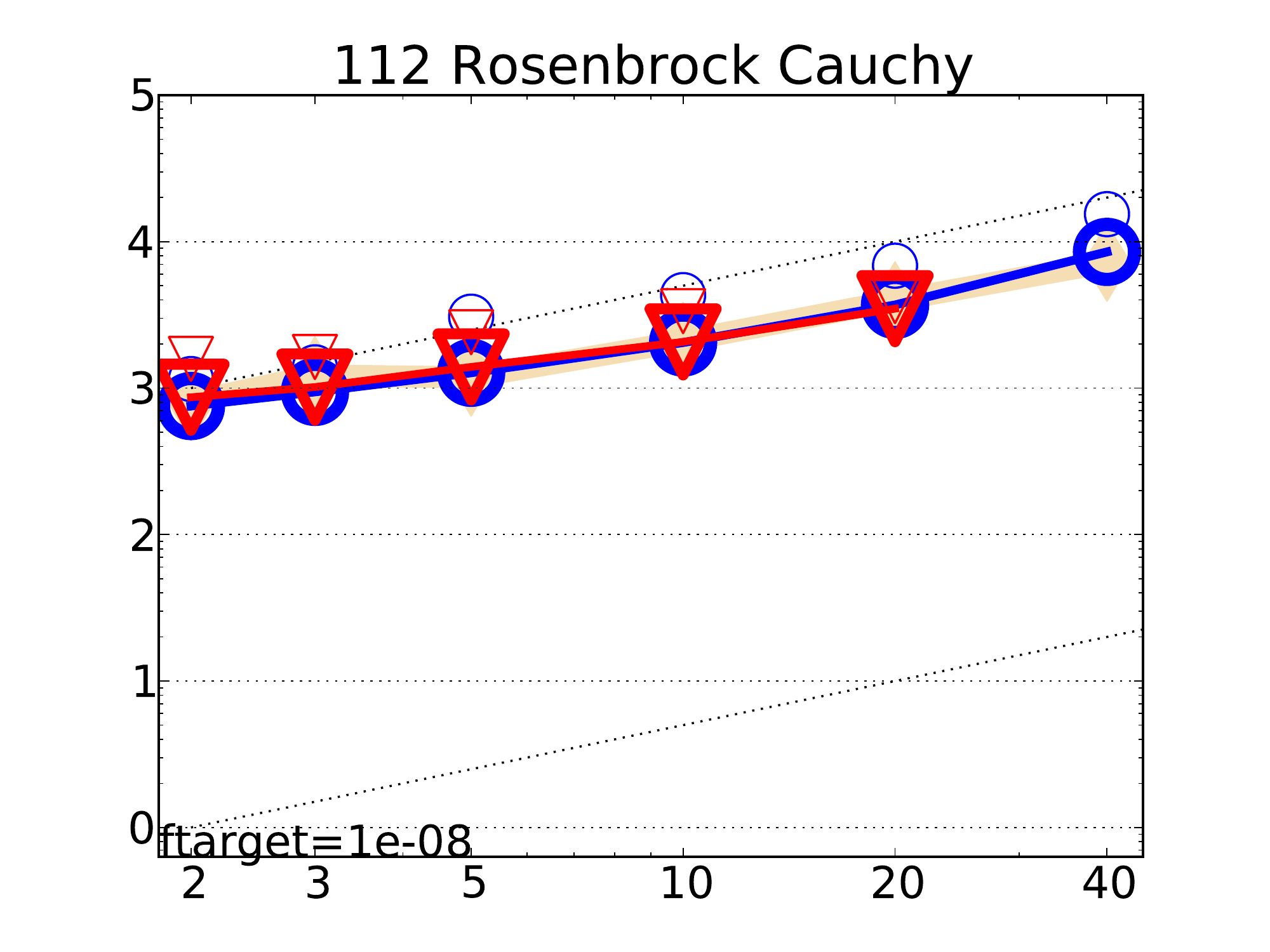}&
\includegraphics[width=0.2\textwidth, trim=20mm 7mm 15mm 3mm, clip]{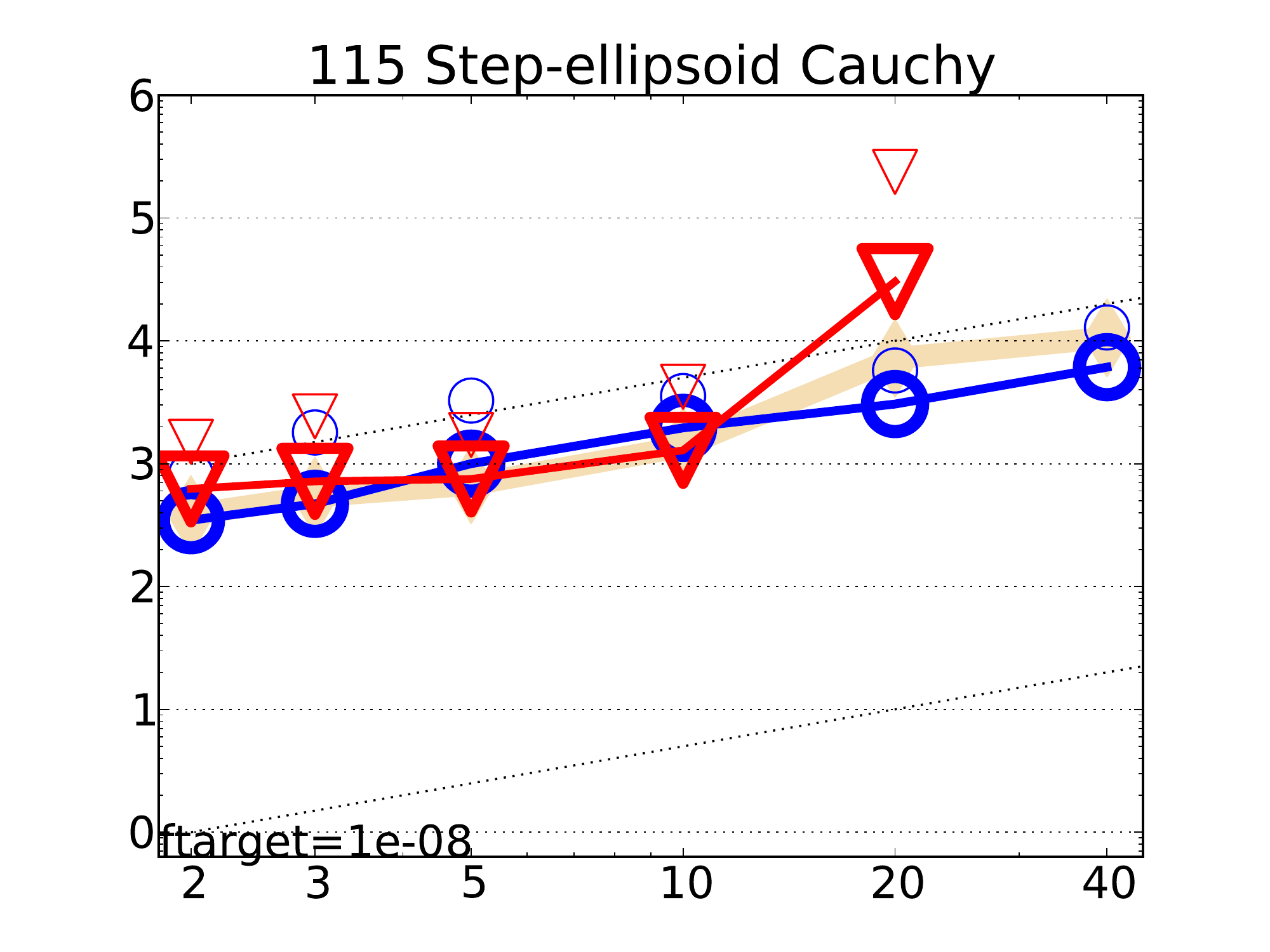}\\
\includegraphics[width=0.2\textwidth, trim=20mm 7mm 15mm 3mm, clip]{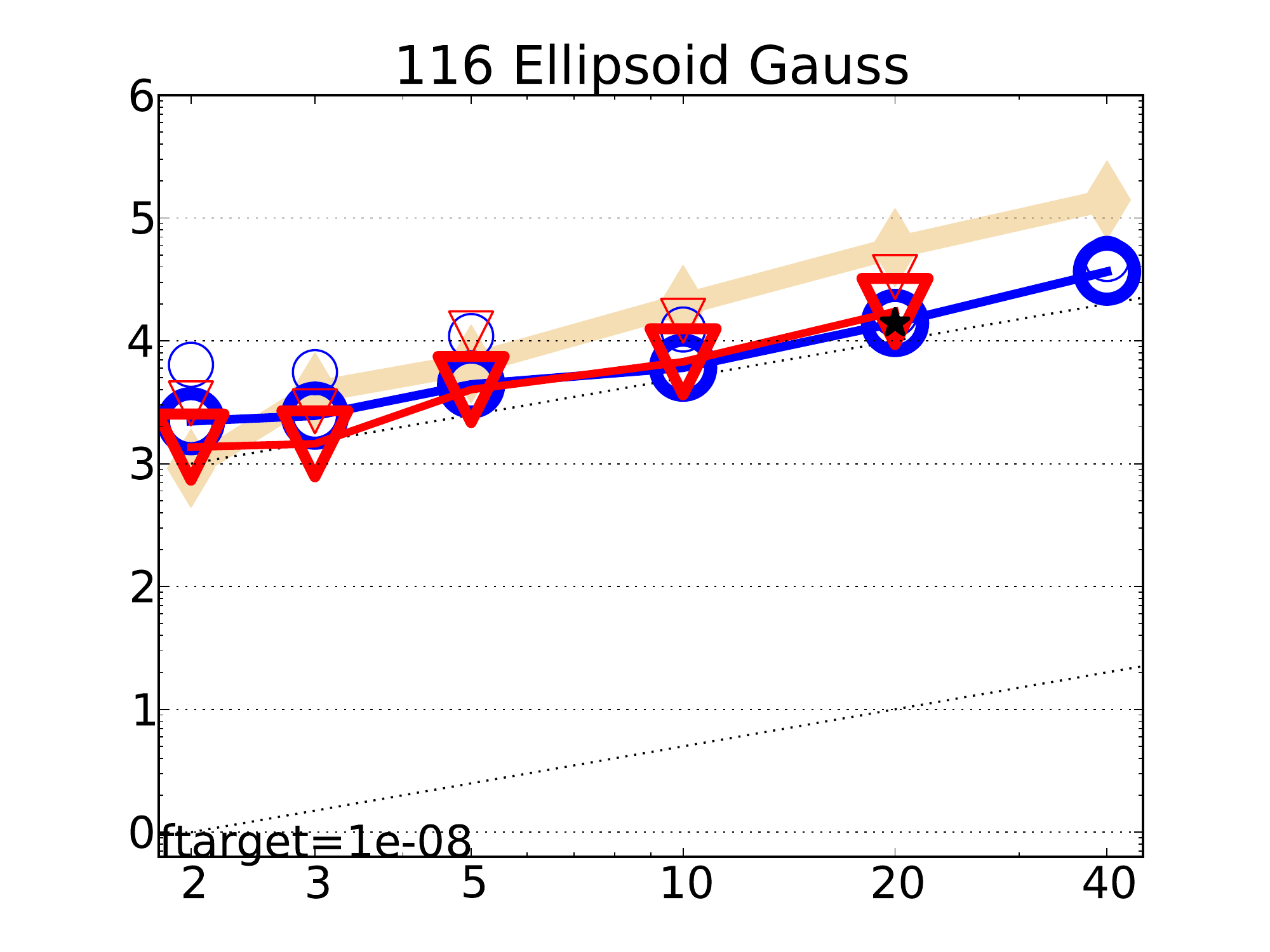}&
\includegraphics[width=0.2\textwidth, trim=20mm 7mm 15mm 3mm, clip]{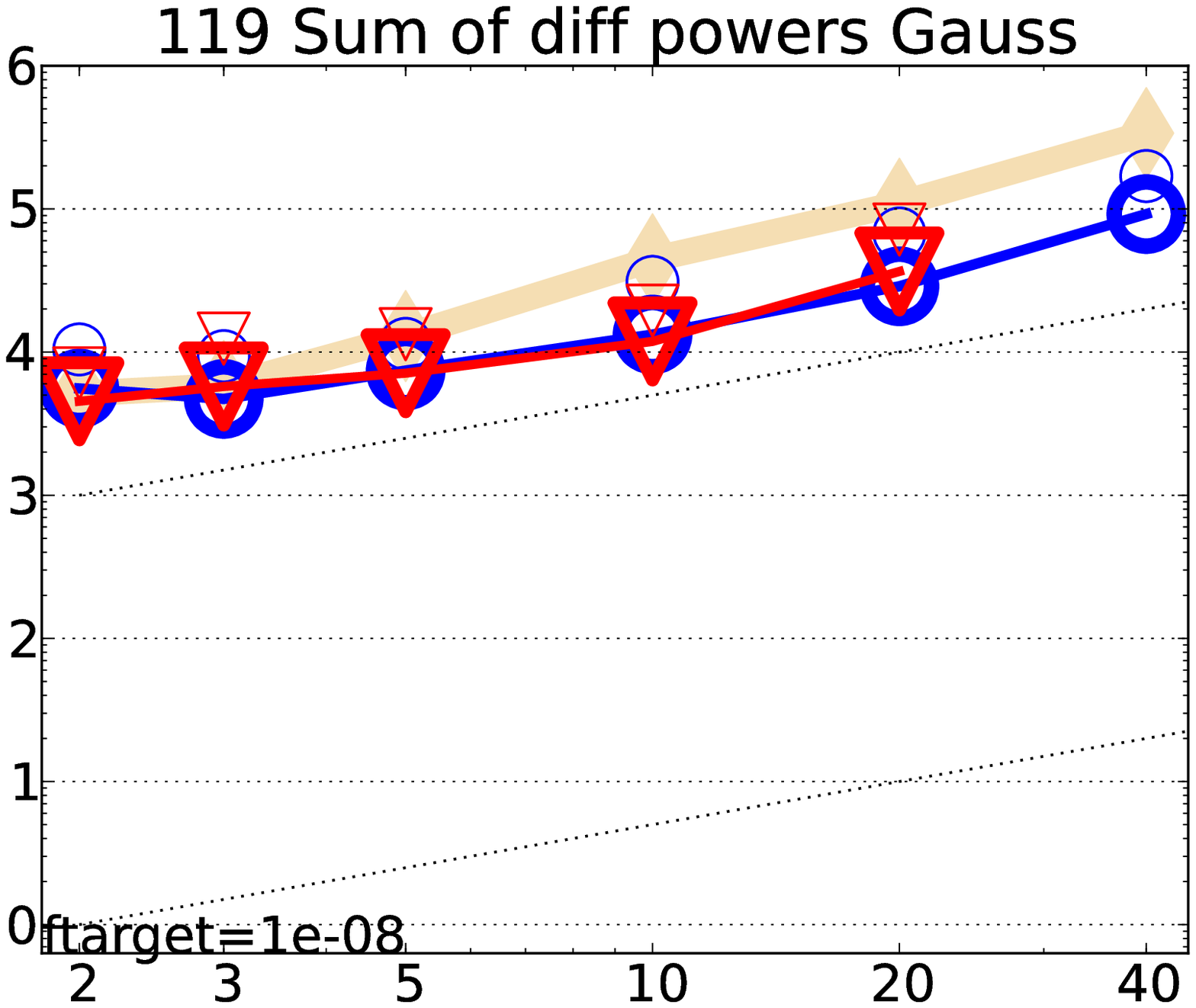}&
\includegraphics[width=0.2\textwidth, trim=20mm 7mm 15mm 3mm, clip]{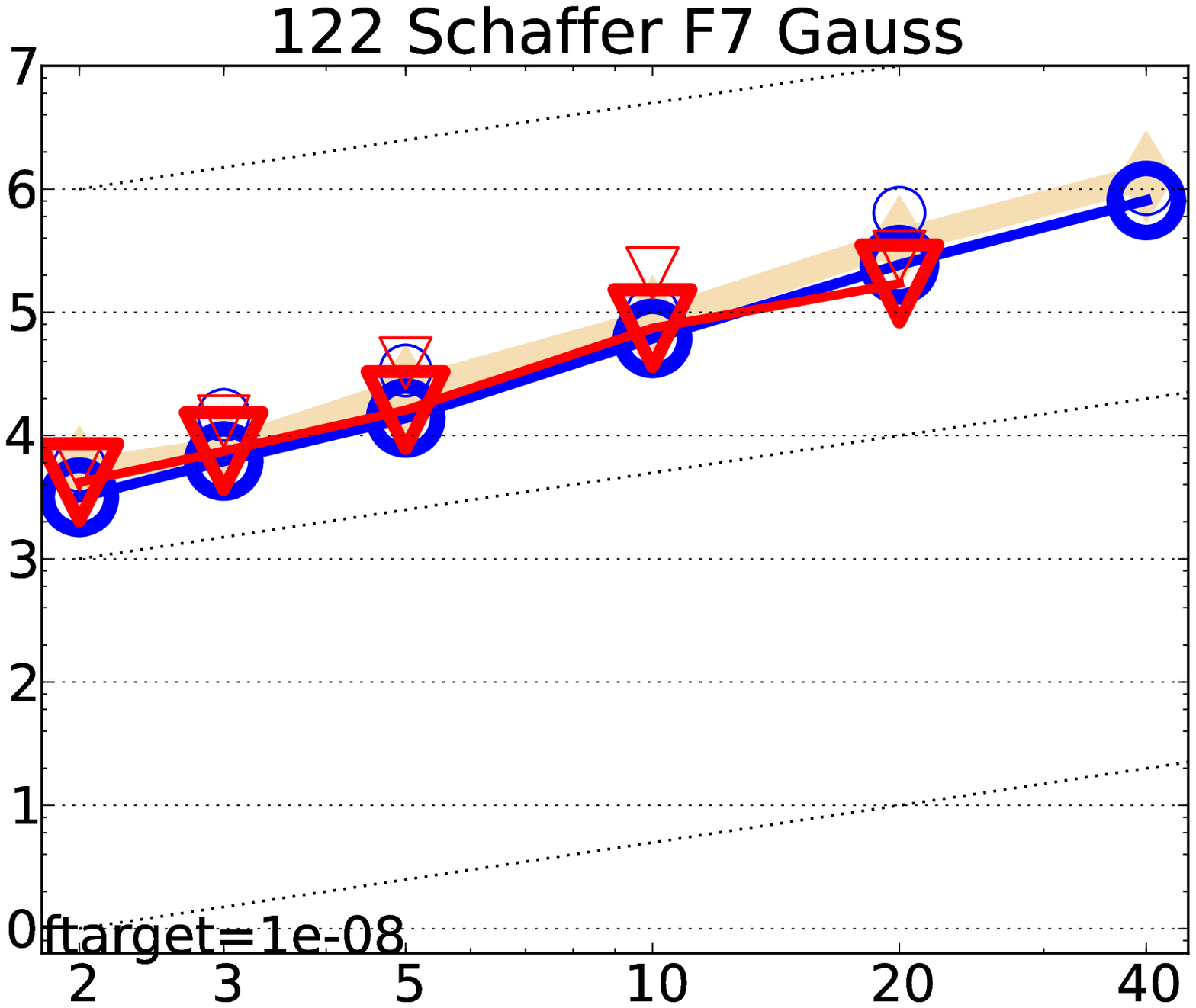}&
\includegraphics[width=0.2\textwidth, trim=20mm 7mm 15mm 3mm, clip]{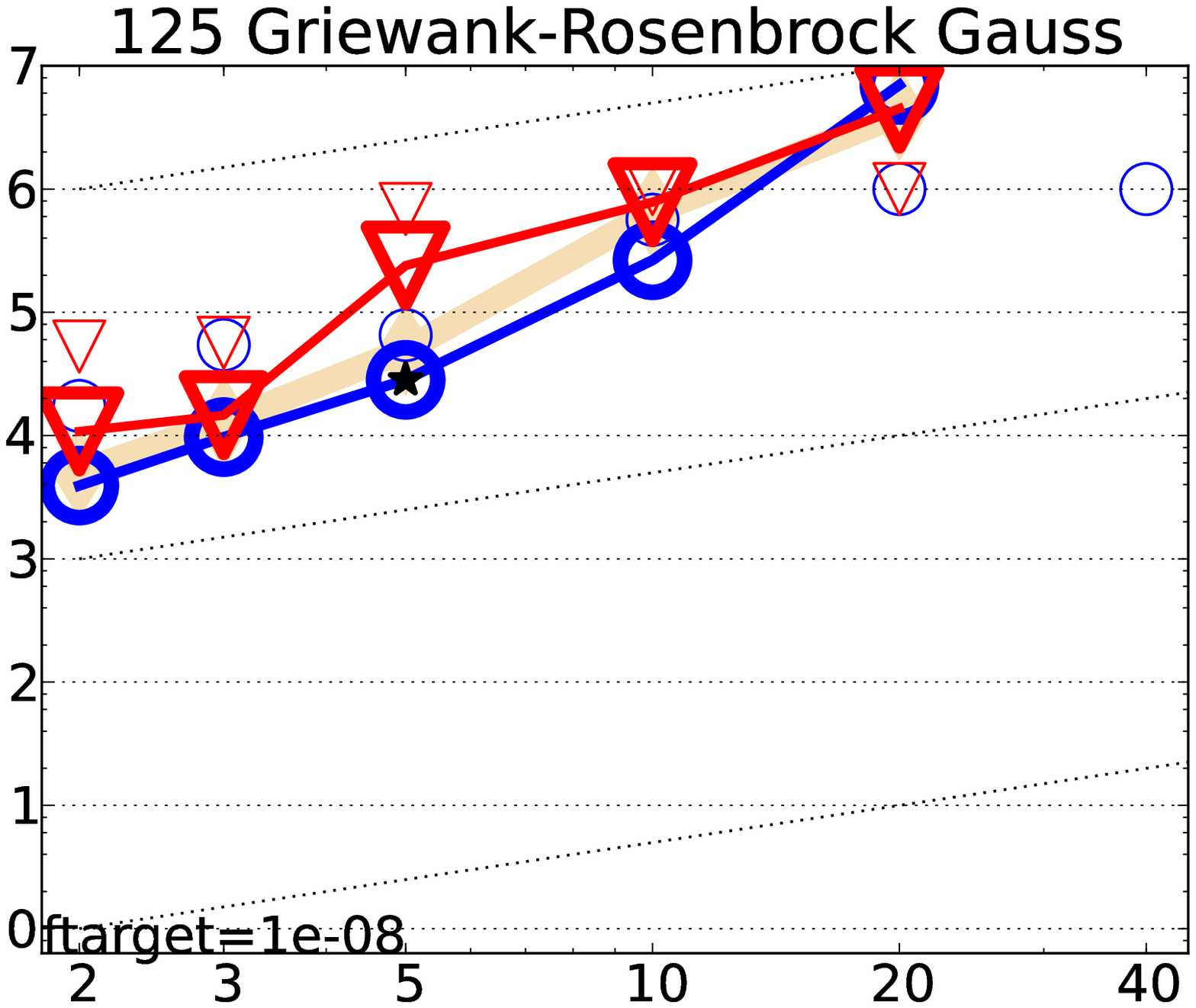}&
\includegraphics[width=0.2\textwidth, trim=20mm 7mm 15mm 3mm, clip]{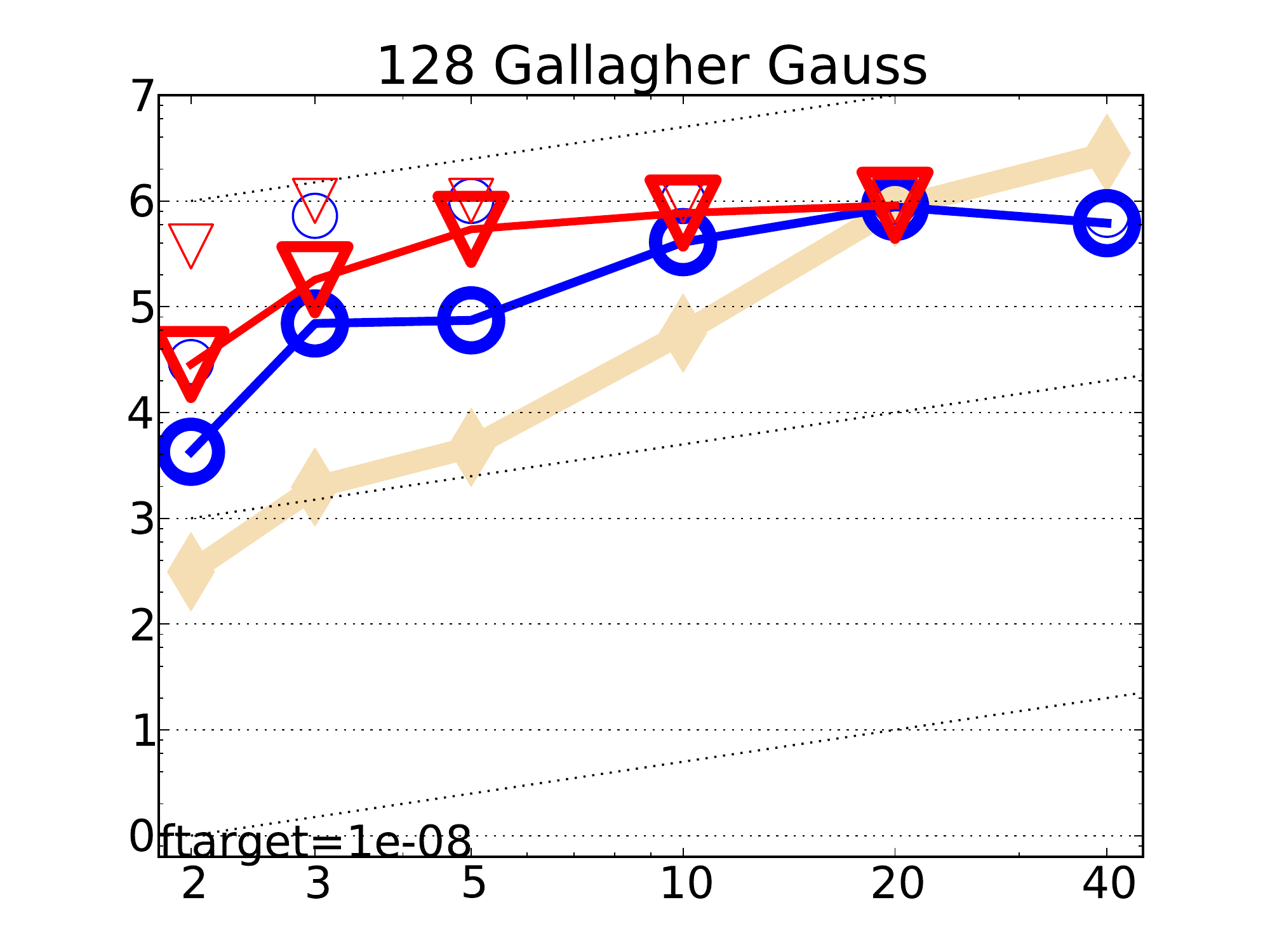}\\
\includegraphics[width=0.2\textwidth, trim=20mm 7mm 15mm 3mm, clip]{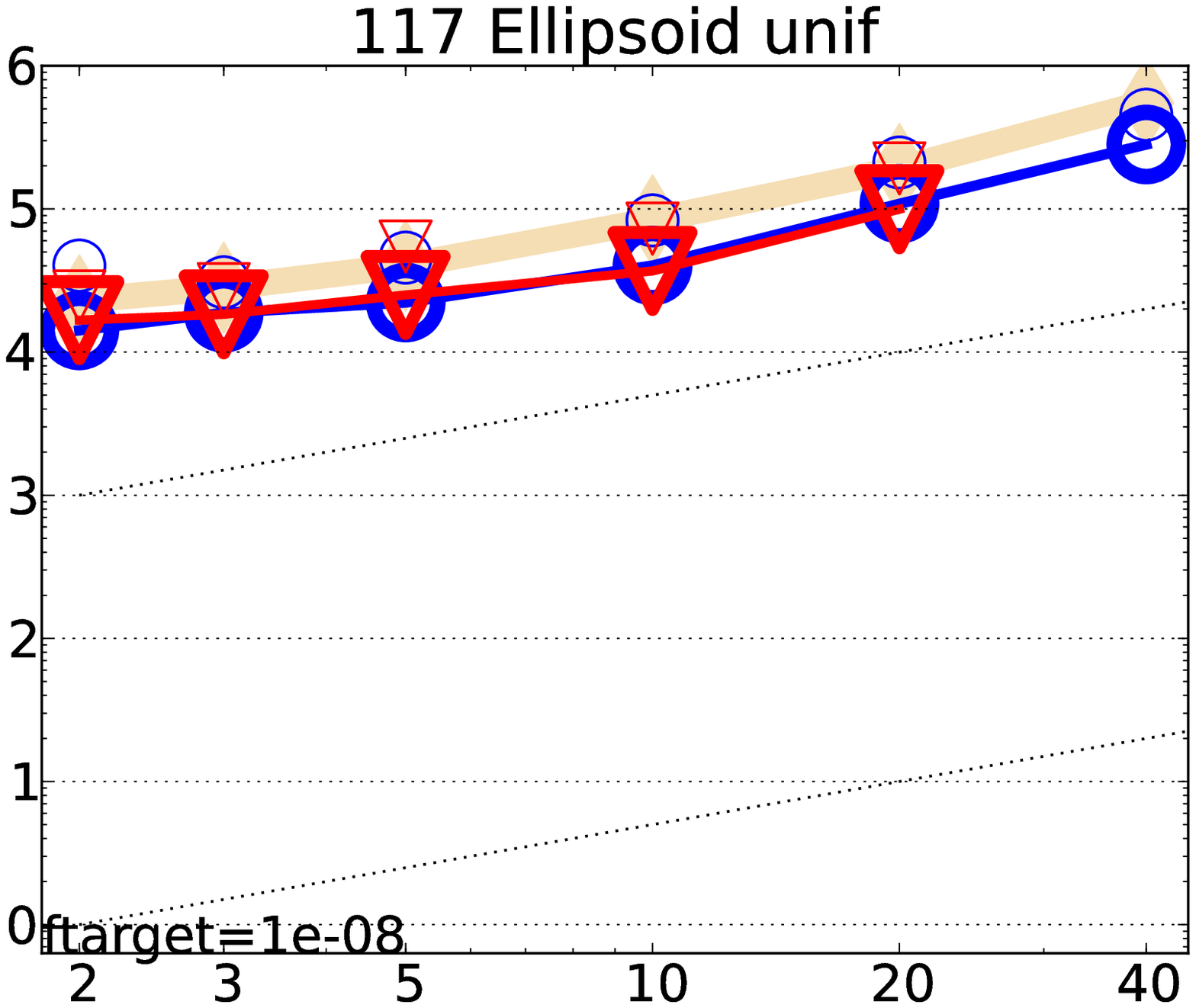}&
\includegraphics[width=0.2\textwidth, trim=20mm 7mm 15mm 3mm, clip]{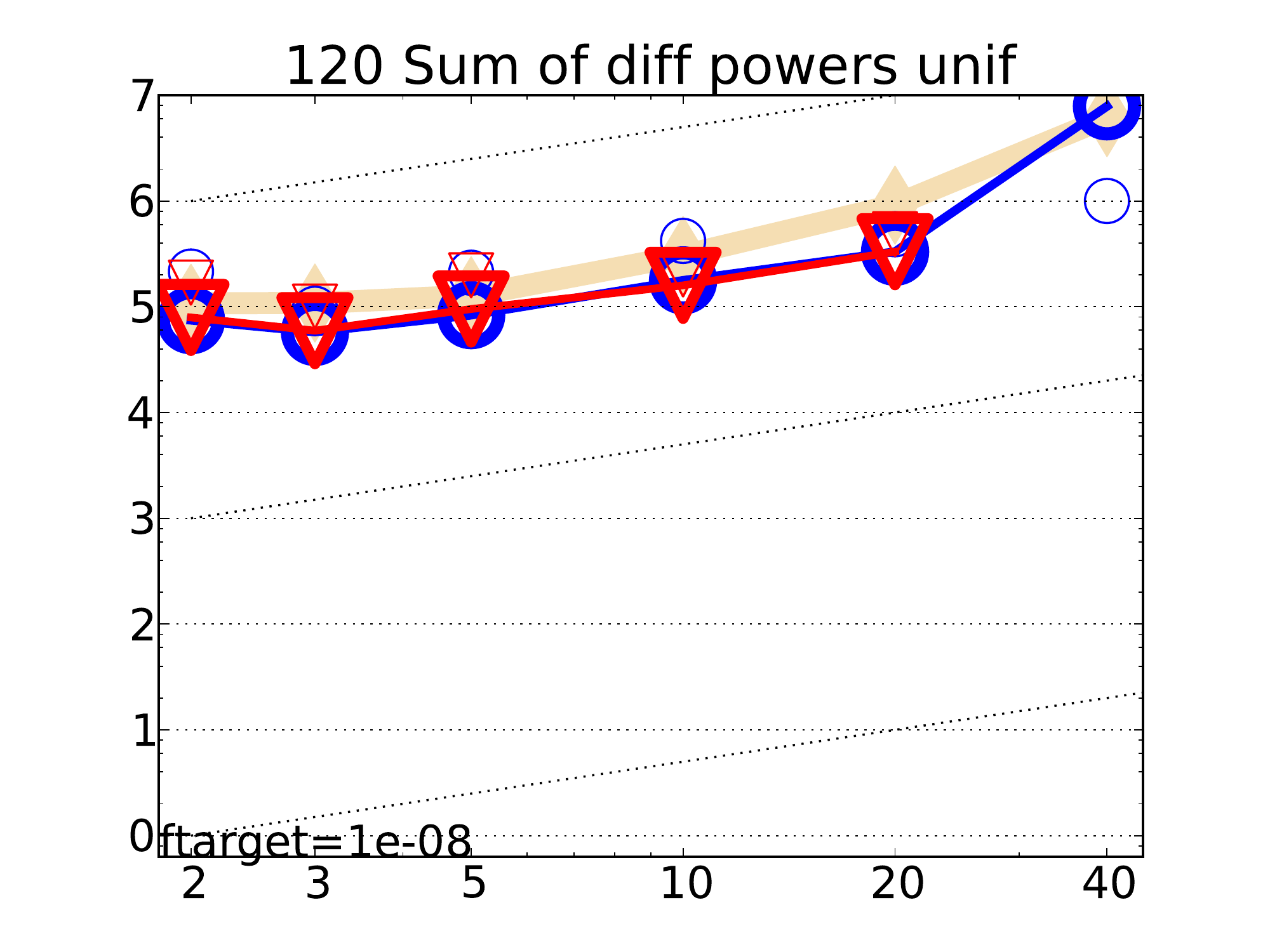}&
\includegraphics[width=0.2\textwidth, trim=20mm 7mm 15mm 3mm, clip]{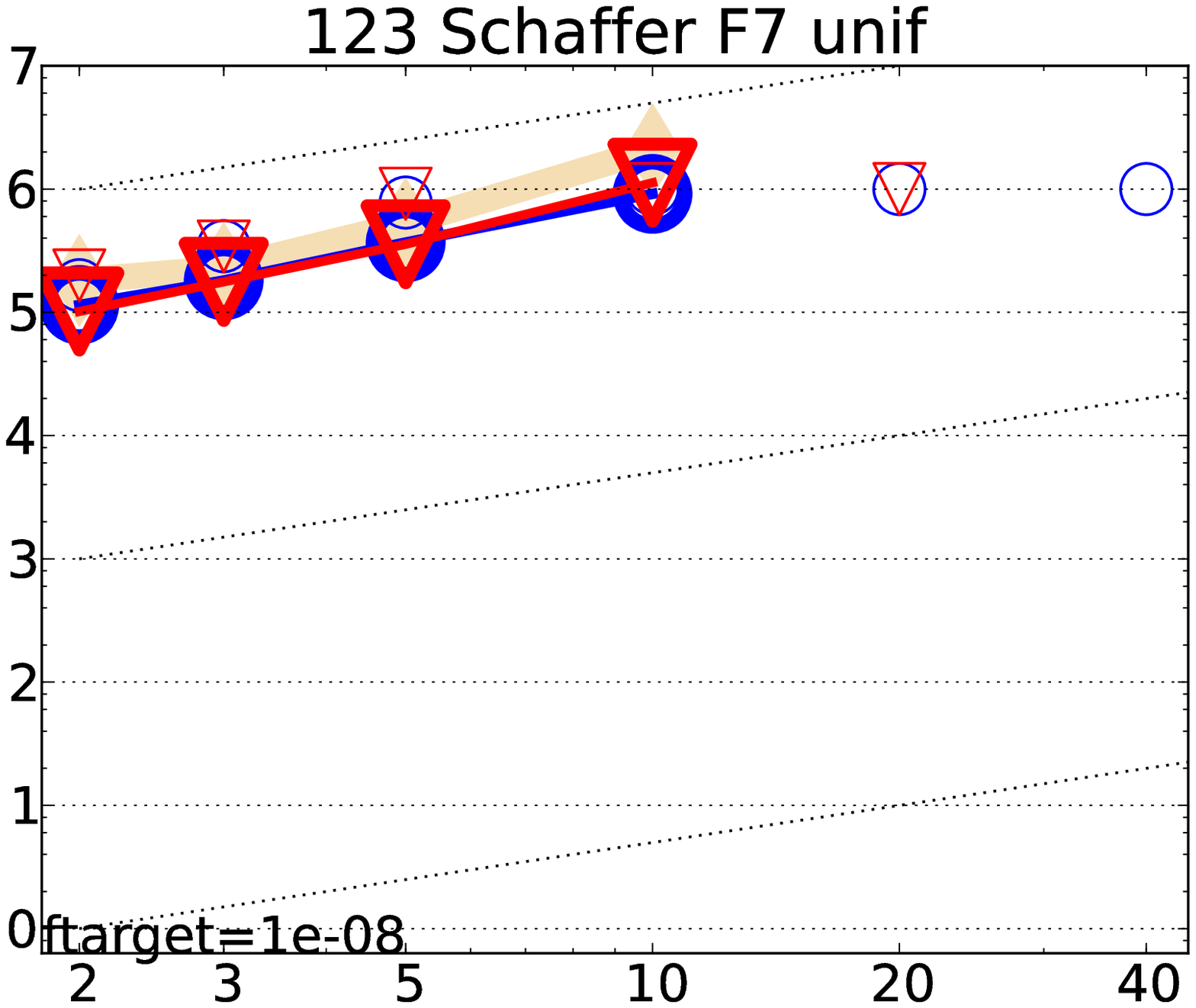}&
\includegraphics[width=0.2\textwidth, trim=20mm 7mm 15mm 3mm, clip]{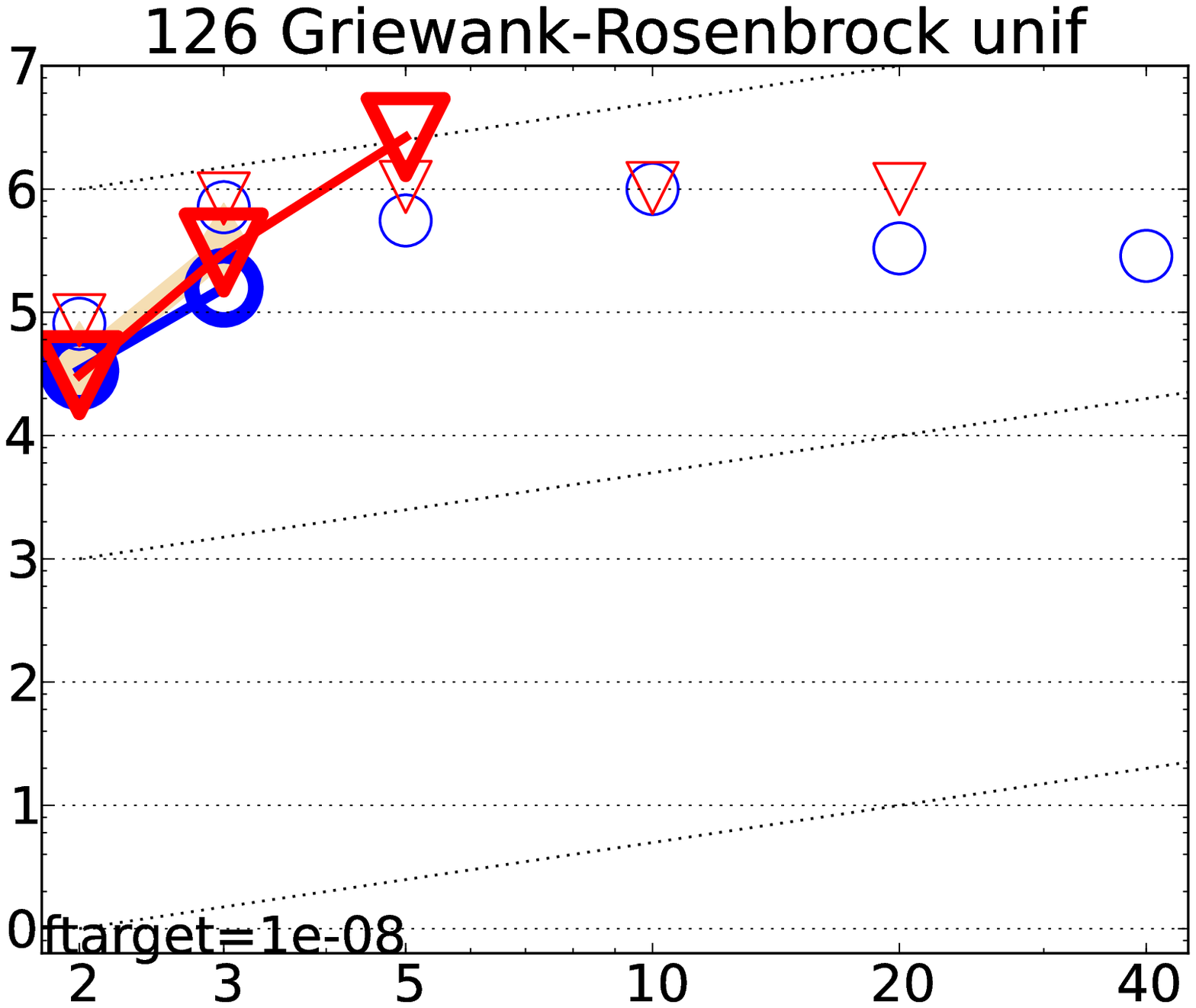}&
\includegraphics[width=0.2\textwidth, trim=20mm 7mm 15mm 3mm, clip]{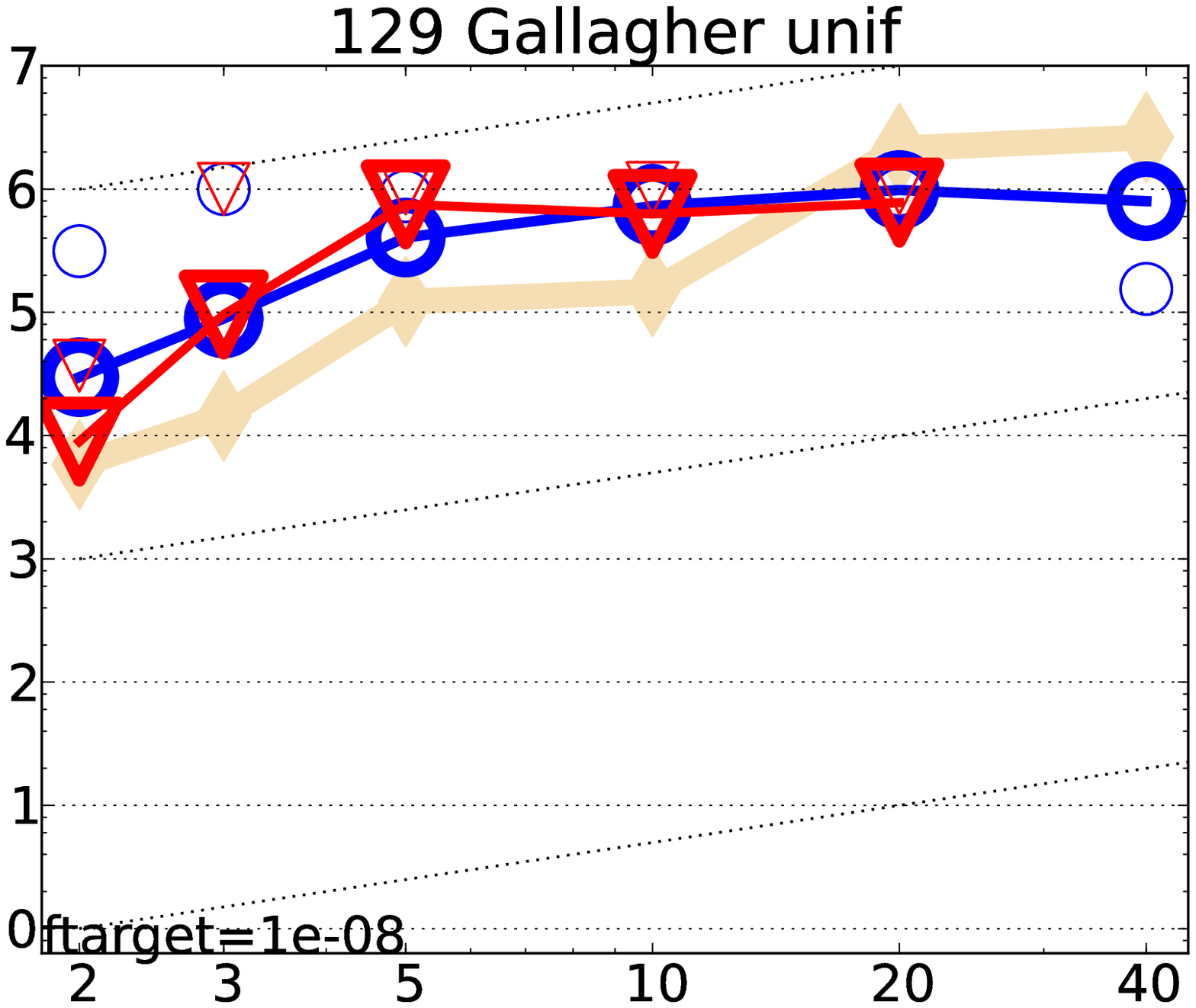}\\
\includegraphics[width=0.2\textwidth, trim=20mm 7mm 15mm 3mm, clip]{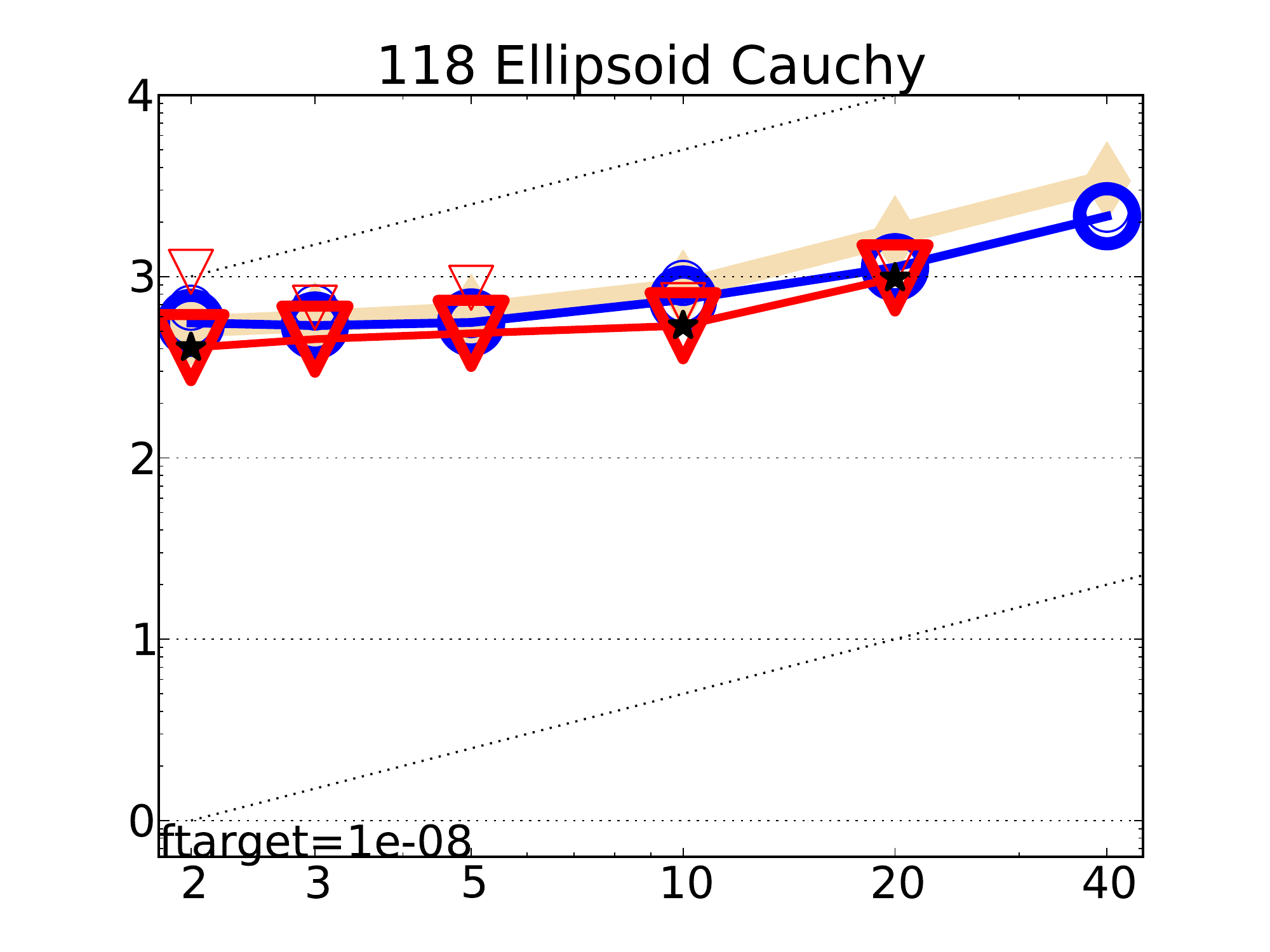}&
\includegraphics[width=0.2\textwidth, trim=20mm 7mm 15mm 3mm, clip]{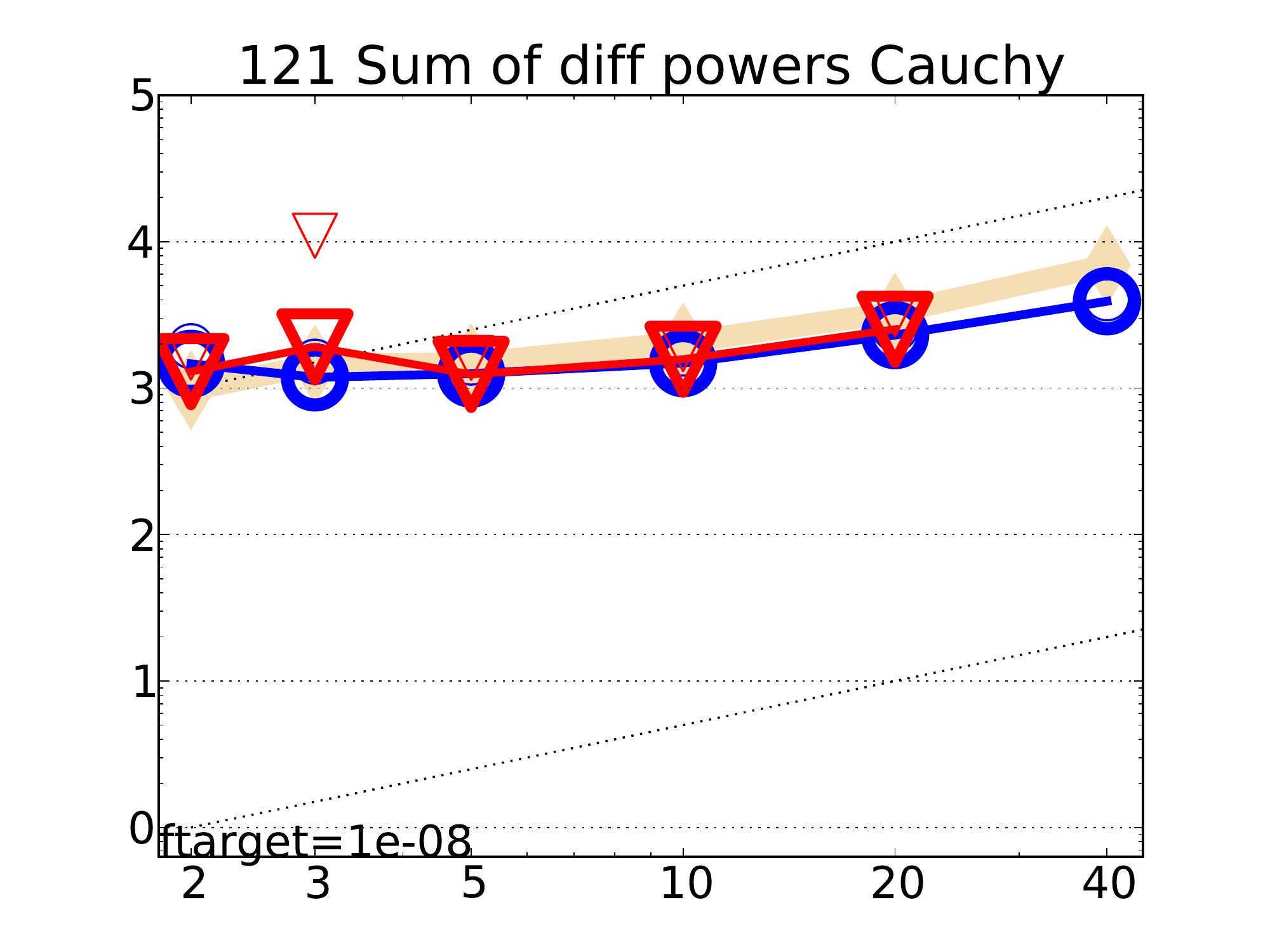}&
\includegraphics[width=0.2\textwidth, trim=20mm 7mm 15mm 3mm, clip]{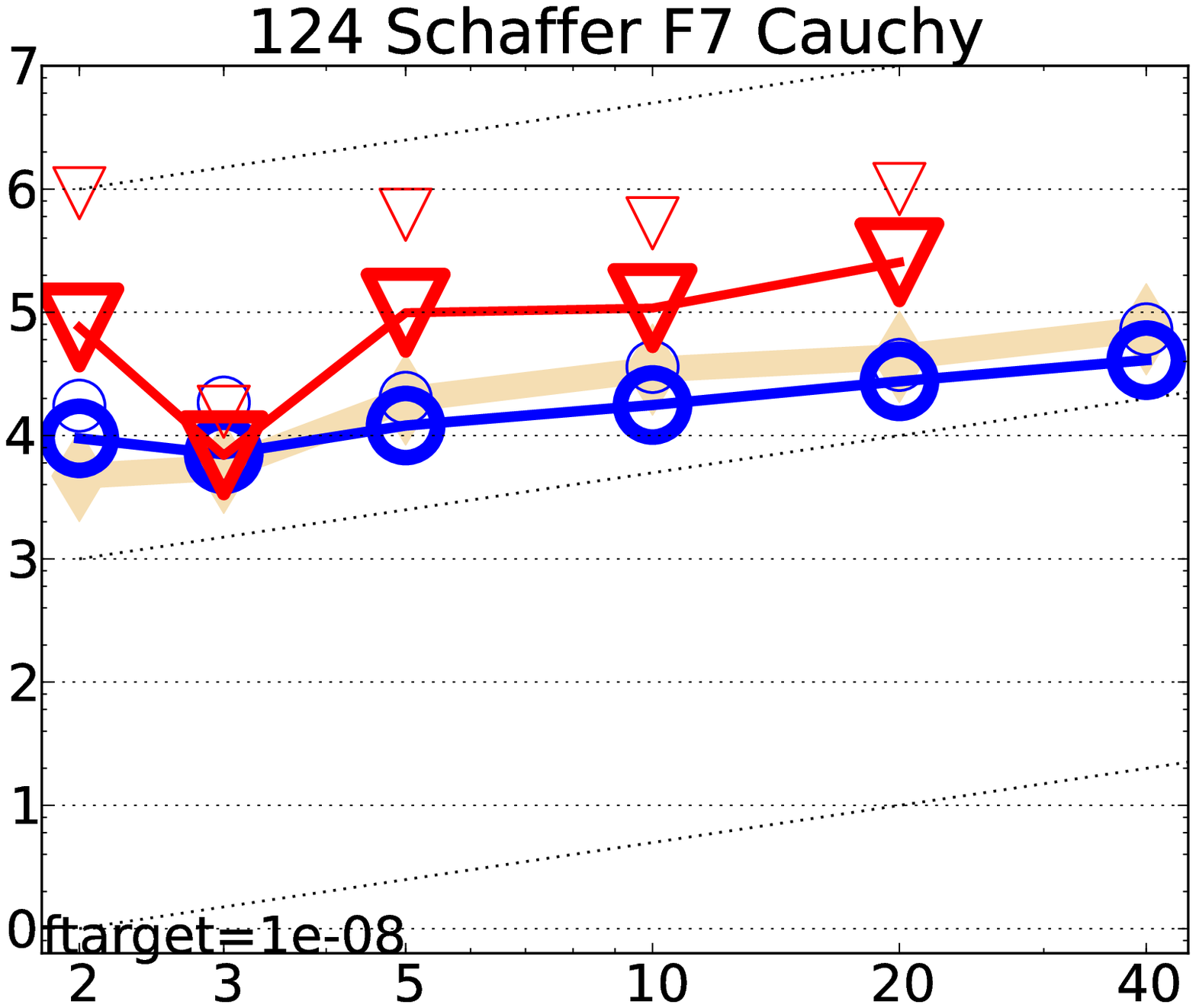}&
\includegraphics[width=0.2\textwidth, trim=20mm 7mm 15mm 3mm, clip]{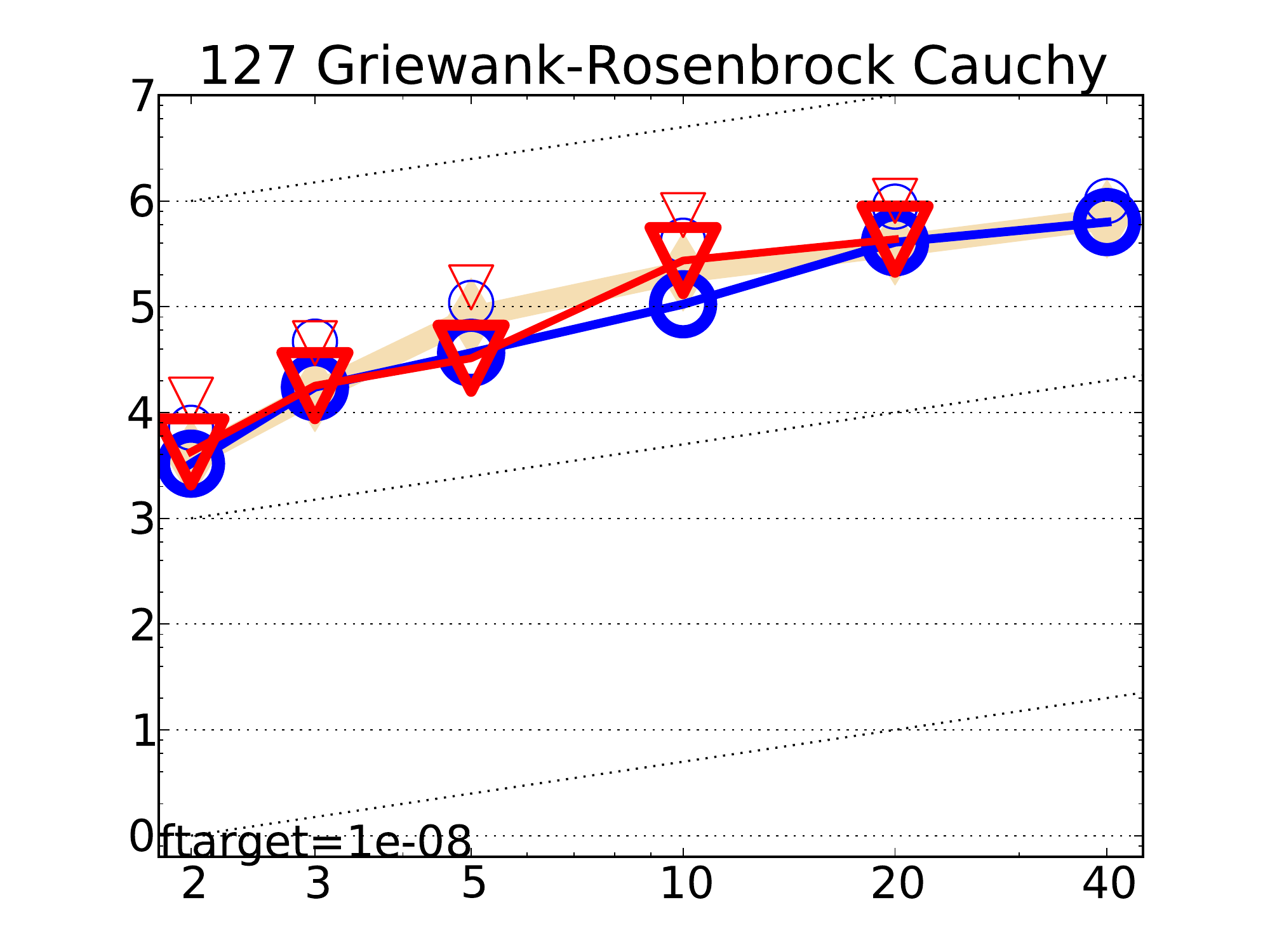}&
\includegraphics[width=0.2\textwidth, trim=20mm 7mm 15mm 3mm, clip]{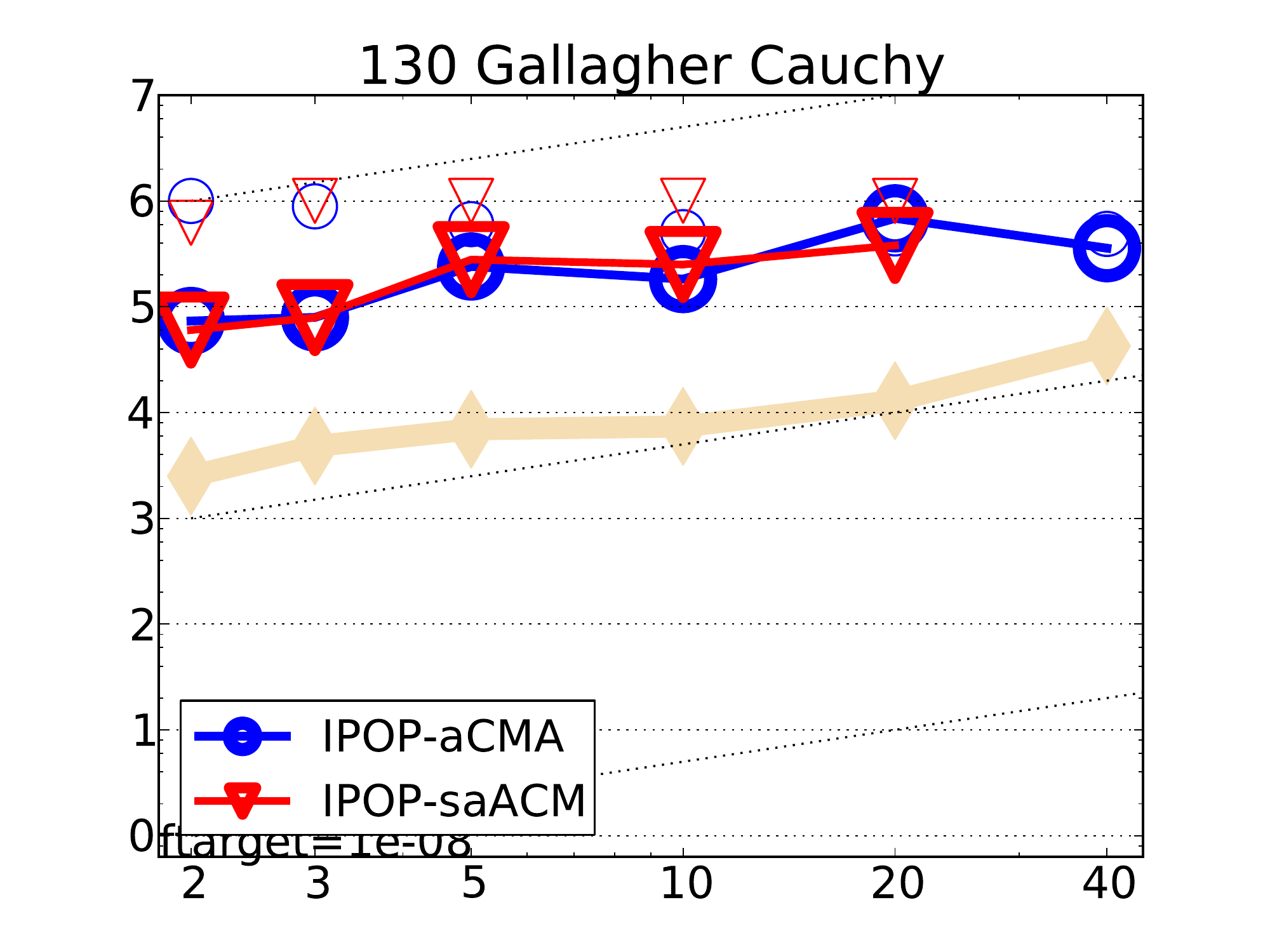}
\end{tabular}
\caption[Expected running time (\ERT) divided by dimension
versus dimension in log-log presentation]{\label{fig:scaling}Expected running
time (\ERT) divided by dimension  for
target function value $10^{-8}$ as $\log_{10}$ values versus dimension. Different symbols
correspond to different algorithms given in legend of $f_{101}$ and $f_{130}$.
Light symbols give the maximum number of function evaluations from all trials divided by the
dimension. Horizontal lines give linear scaling,
the slanted dotted lines give quadratic scaling.
\input{\bbobdatapath ppfigs}
}
\end{figure*}

\begin{figure*}
 \begin{tabular}{@{}c@{}c@{}}
all functions & moderate noise \\
  \input{\bbobdatapath pprldmany_05D_nzall}%
  \includegraphics[width=0.4135\textwidth,trim=0mm 0mm 34mm 10mm, clip]{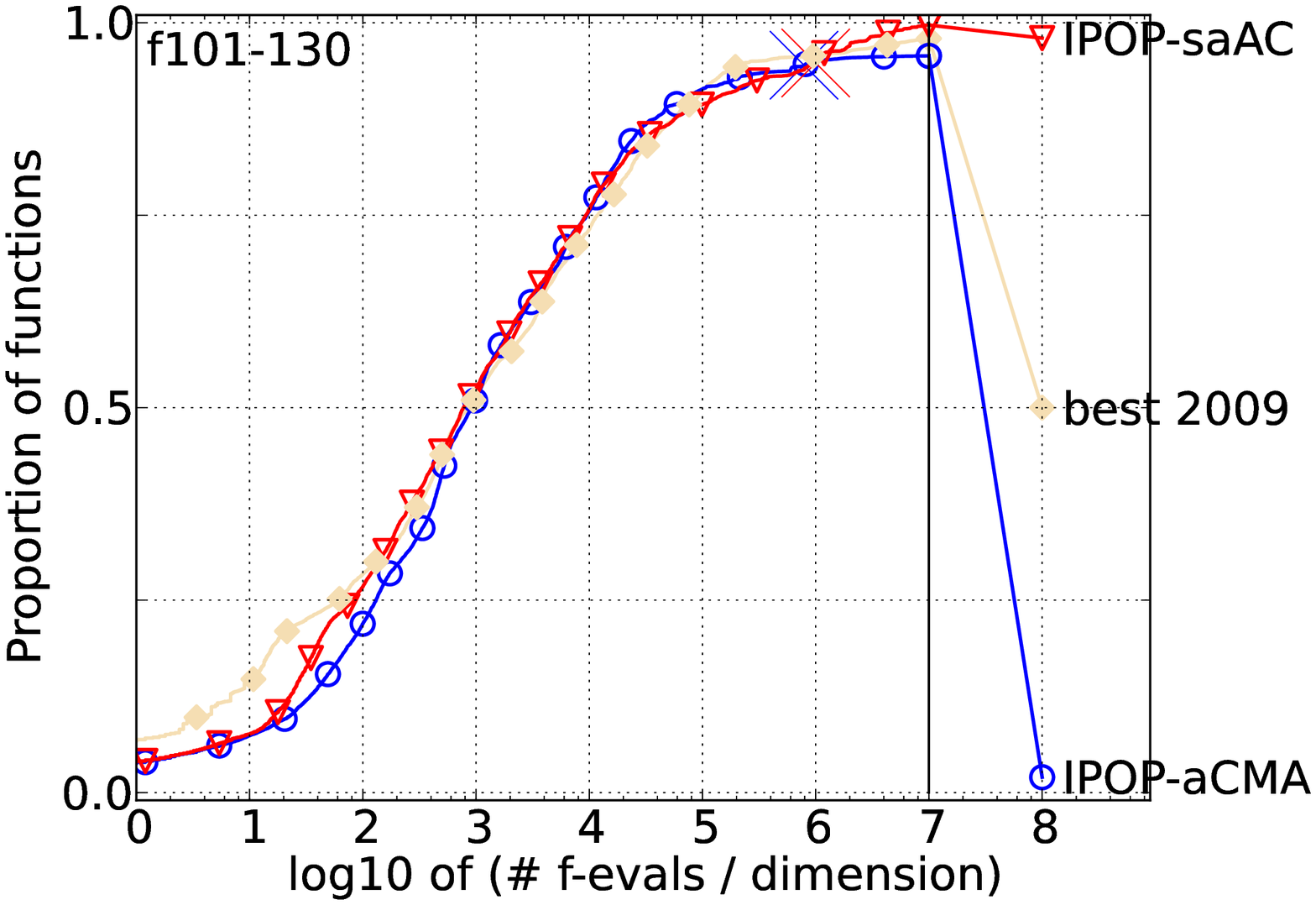}%
  \raisebox{.037\textwidth}{\parbox[b][.3\textwidth]{.0868\textwidth}{\begin{scriptsize}
    \perfprofsidepanel 
  \end{scriptsize}}}
 & 
  \input{\bbobdatapath pprldmany_05D_nzmod}%
  \includegraphics[width=0.4135\textwidth,trim=0mm 0mm 34mm 10mm, clip]{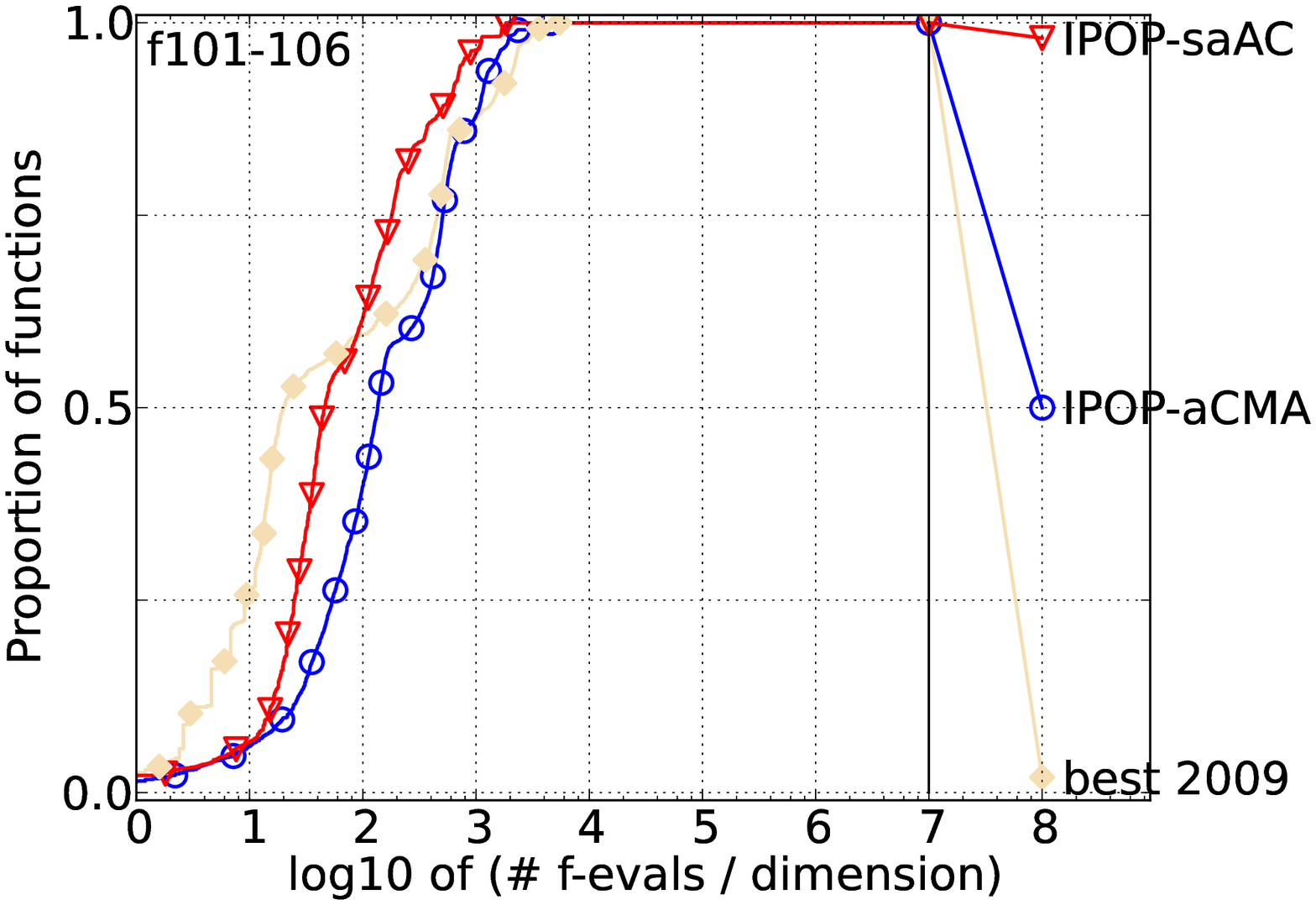}%
  \raisebox{.037\textwidth}{\parbox[b][.3\textwidth]{.0868\textwidth}{\begin{scriptsize}
    \perfprofsidepanel 
  \end{scriptsize}}}
 \\
severe noise & severe noise multimod.\\
  \input{\bbobdatapath pprldmany_05D_nzsev}%
  \includegraphics[width=0.4135\textwidth,trim=0mm 0mm 34mm 10mm, clip]{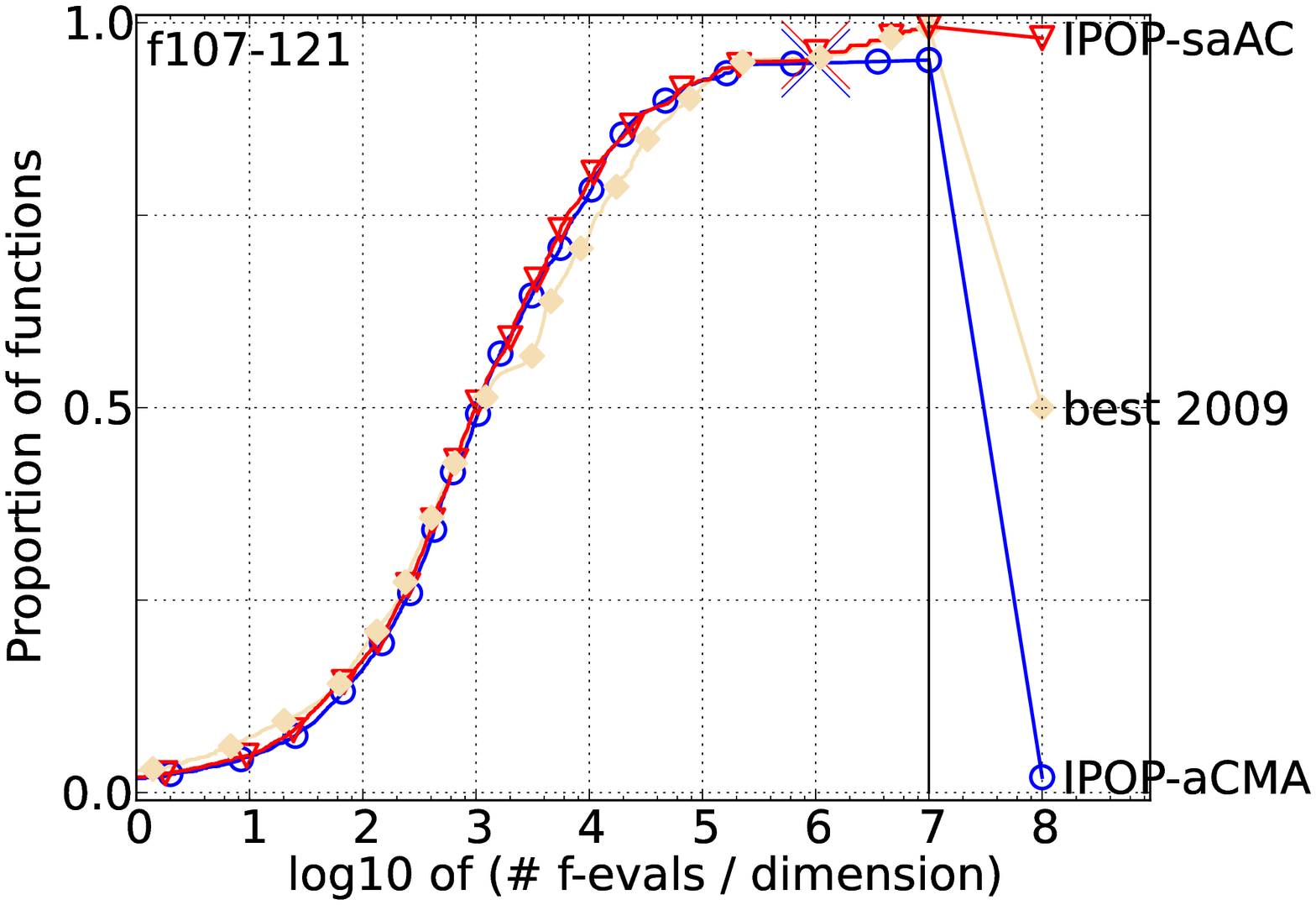}%
  \raisebox{.037\textwidth}{\parbox[b][.3\textwidth]{.0868\textwidth}{\begin{scriptsize}
    \perfprofsidepanel 
  \end{scriptsize}}}
 & 
  \input{\bbobdatapath pprldmany_05D_nzsmm}%
  \includegraphics[width=0.4135\textwidth,trim=0mm 0mm 34mm 10mm, clip]{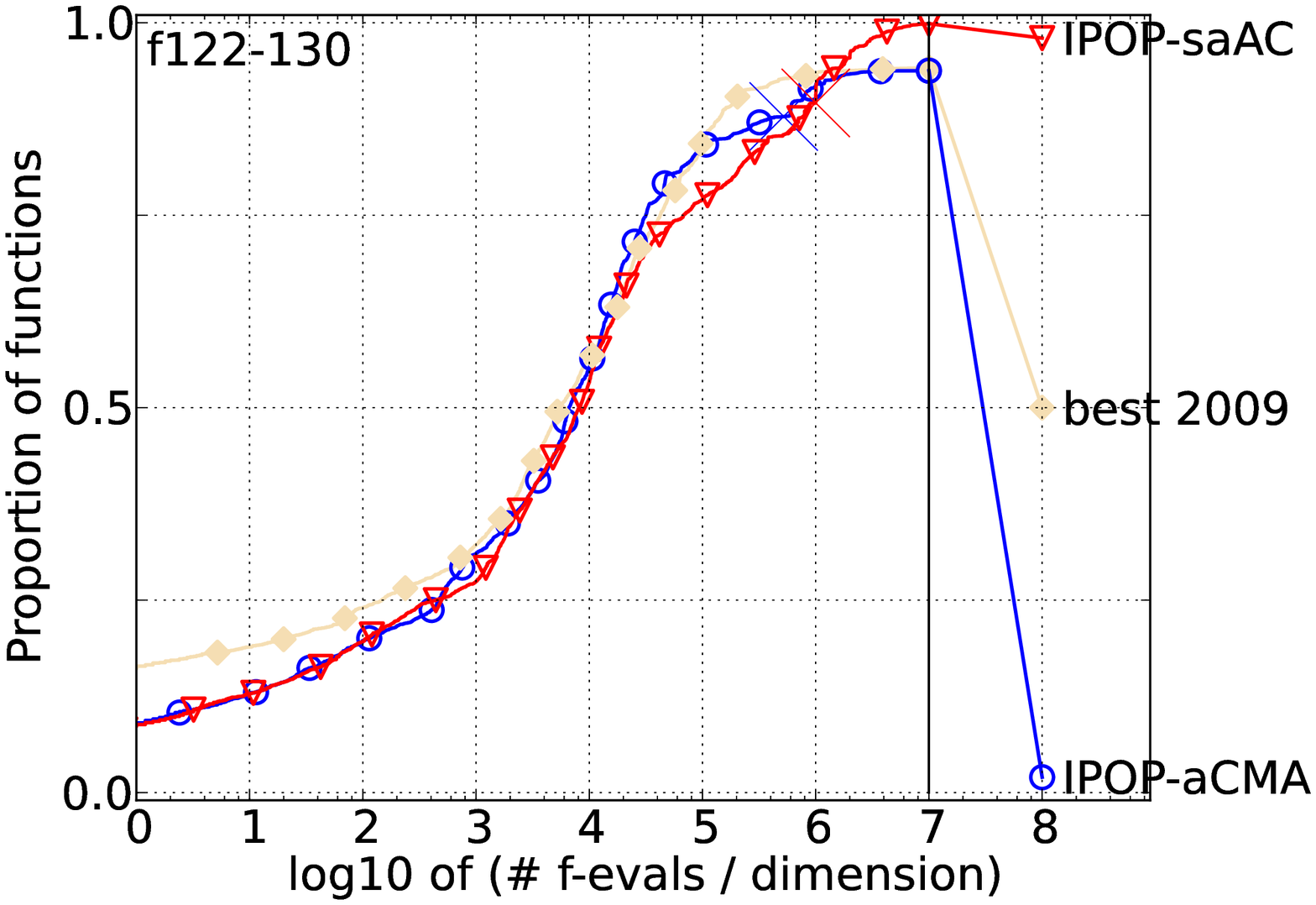}%
  \raisebox{.037\textwidth}{\parbox[b][.3\textwidth]{.0868\textwidth}{\begin{scriptsize}
    \perfprofsidepanel 
  \end{scriptsize}}}

 \end{tabular}
\caption{
\label{fig:ECDFs05D}
Bootstrapped empirical cumulative distribution of 
the number of objective function evaluations
divided by dimension (FEvals/D) for 50 targets in
$10^{[-8..2]}$ for all functions and subgroups in 5-D. The ``best 2009'' line
corresponds to the best \ERT\ observed during BBOB 2009 for each single target. 
}
\end{figure*}

\begin{figure*}
 \begin{tabular}{@{}c@{}c@{}}
all functions & moderate noise \\
  \input{\bbobdatapath pprldmany_20D_nzall}%
  \includegraphics[width=0.4135\textwidth,trim=0mm 0mm 34mm 10mm, clip]{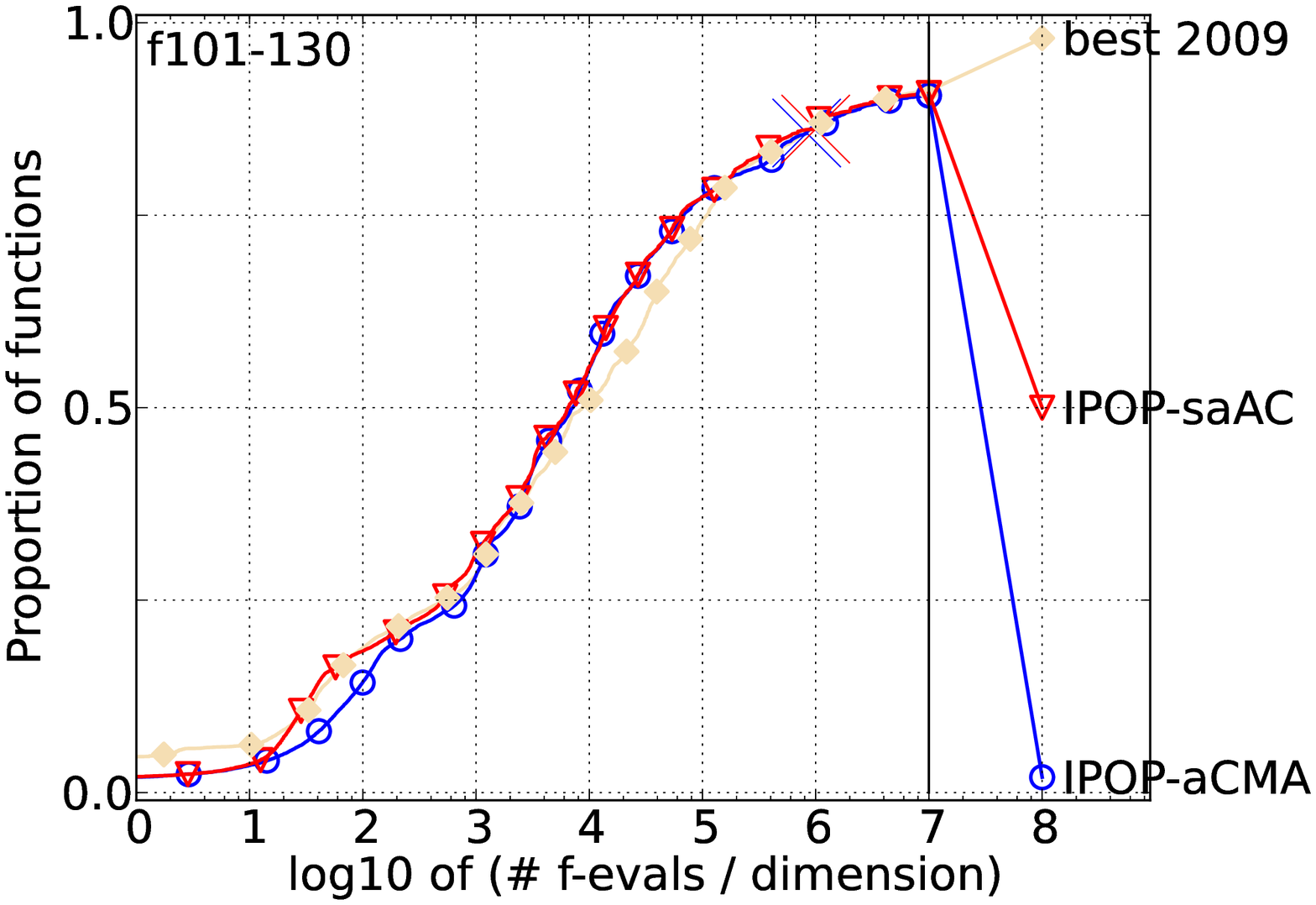}%
  \raisebox{.037\textwidth}{\parbox[b][.3\textwidth]{.0868\textwidth}{\begin{scriptsize}
    \perfprofsidepanel 
  \end{scriptsize}}}
 & 
  \input{\bbobdatapath pprldmany_20D_nzmod}%
  \includegraphics[width=0.4135\textwidth,trim=0mm 0mm 34mm 10mm, clip]{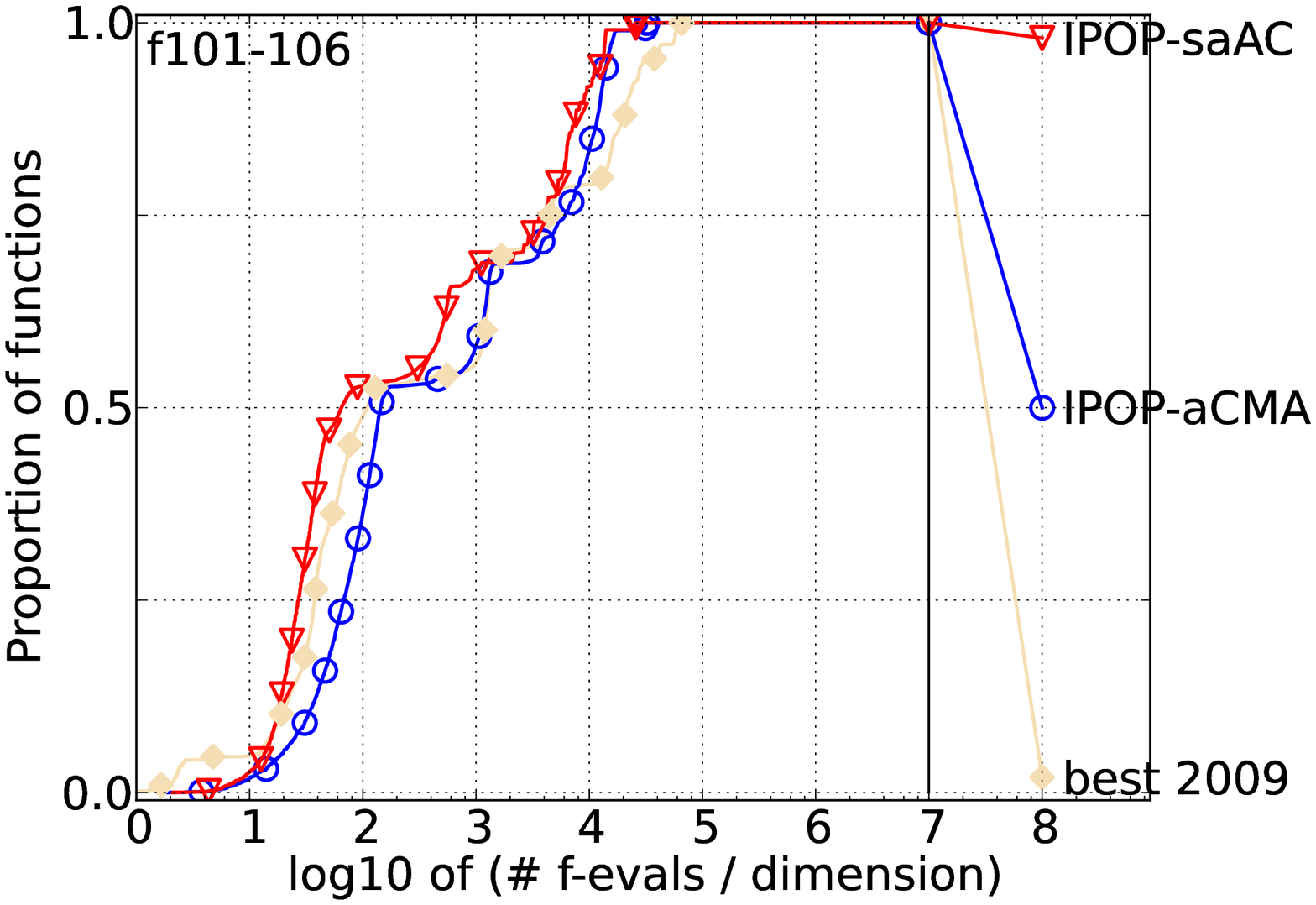}%
  \raisebox{.037\textwidth}{\parbox[b][.3\textwidth]{.0868\textwidth}{\begin{scriptsize}
    \perfprofsidepanel 
  \end{scriptsize}}}
 \\
severe noise & severe noise multimod.\\
  \input{\bbobdatapath pprldmany_20D_nzsev}%
  \includegraphics[width=0.4135\textwidth,trim=0mm 0mm 34mm 10mm, clip]{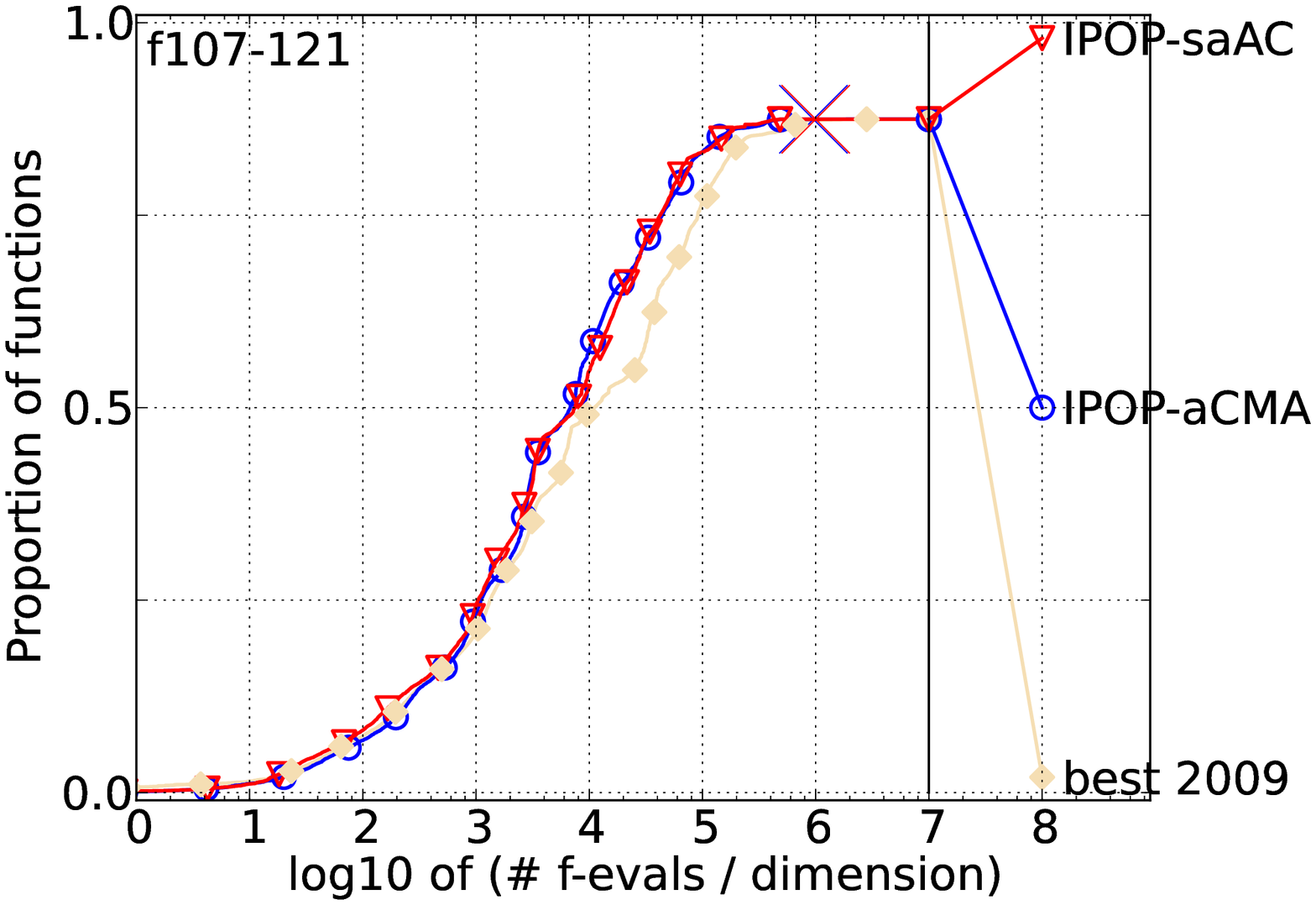}%
  \raisebox{.037\textwidth}{\parbox[b][.3\textwidth]{.0868\textwidth}{\begin{scriptsize}
    \perfprofsidepanel 
  \end{scriptsize}}}
 & 
  \input{\bbobdatapath pprldmany_20D_nzsmm}%
  \includegraphics[width=0.4135\textwidth,trim=0mm 0mm 34mm 10mm, clip]{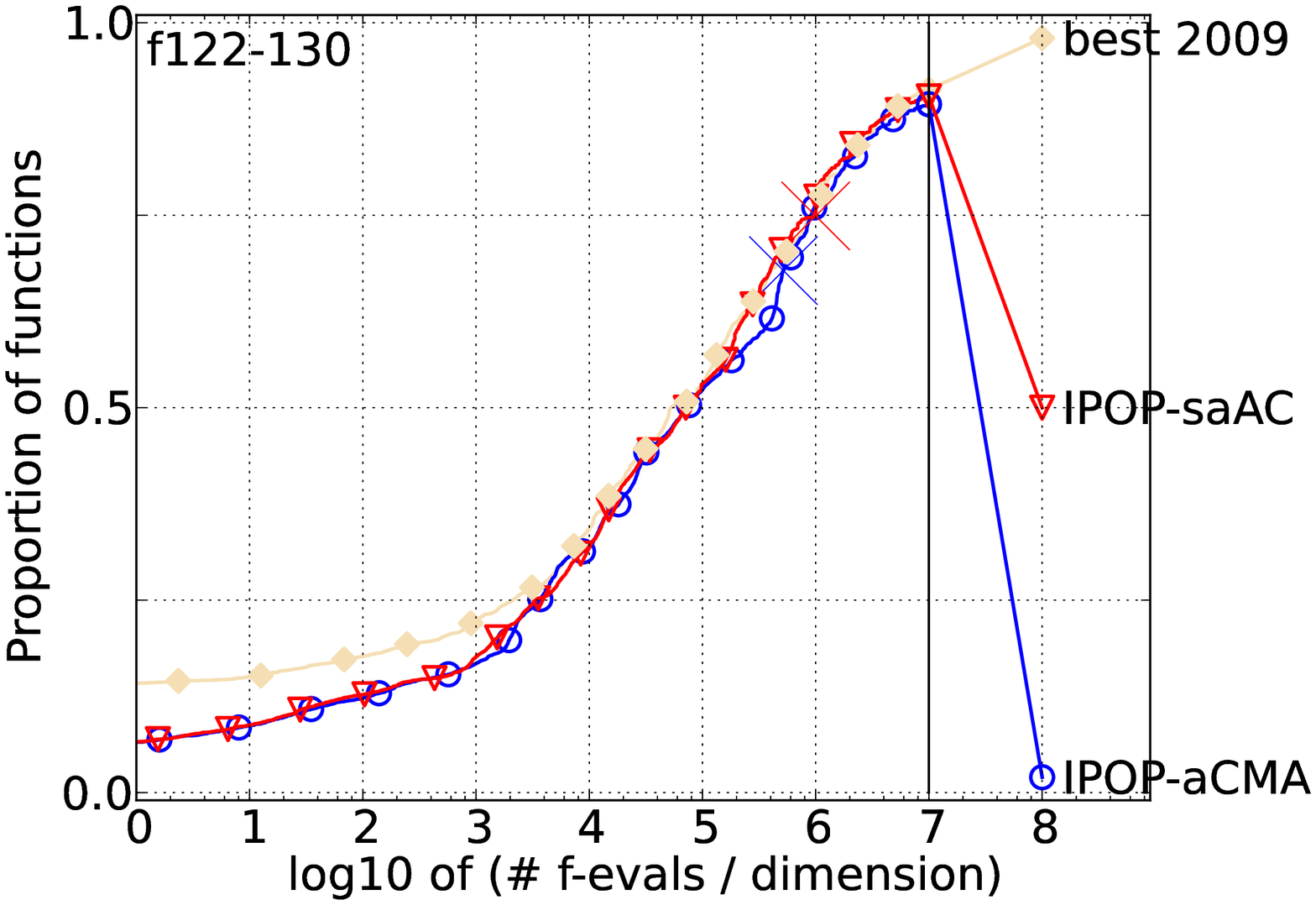}%
  \raisebox{.037\textwidth}{\parbox[b][.3\textwidth]{.0868\textwidth}{\begin{scriptsize}
    \perfprofsidepanel 
  \end{scriptsize}}}

 \end{tabular}
\caption{
\label{fig:ECDFs20D}
Bootstrapped empirical cumulative distribution of 
the number of objective function evaluations
divided by dimension (FEvals/D) for 50 targets in
$10^{[-8..2]}$ for all functions and subgroups in 20-D. The ``best 2009'' line
corresponds to the best \ERT\ observed during BBOB 2009 for each single target. 
}
\end{figure*}

Results from experiments according to \cite{hansen2012exp} on the benchmark
functions given in \cite{wp200902_2010,hansen2012noi} are presented in 
Figures~\ref{fig:ECDFs05D} and \ref{fig:ECDFs20D}, and Figure~\ref{fig:scaling}.
The \textbf{expected running time (\ERT)}, used in the figures and table,
depends on a given target function value, $\ftarget=\fopt+\Df$, and is computed
over all relevant trials (on the first 15 instances) as the number of function evaluations executed during
each trial while the best function value did not reach \ftarget, summed over
all trials and divided by the number of trials that actually reached \ftarget\
\cite{hansen2012exp,price1997dev}.
\textbf{Statistical significance} is tested with the rank-sum test for a given
target $\Delta\ftarget$ using, for each trial, either the number of needed
function evaluations to reach $\Delta\ftarget$ (inverted and multiplied by
$-1$), or, if the target was not reached, the best $\Df$-value achieved,
measured only up to the smallest number of overall function evaluations for any
unsuccessful trial under consideration if available.
Tables~\ref{tab:ERTs5} and~\ref{tab:ERTs20} give the Expected Running Time (\ERT) for targets
$10^{1,\,-1,\,-3,\,-5,\,-7}$ divided by the best \ERT\ obtained during
BBOB-2009 (given in the \ERT$_{\text{best}}$ row), respectively in 5-D
and 20-D.
Bold entries correspond to the best (or 3-best if there are more than 3
algorithms) values.
The median number of conducted function evaluations is additionally given in
\textit{italics}, if $\ERT(10^{-7}) =\infty$.
\#succ is the number of trials that reached the final target $\fopt + 10^{-8}$.
Entries with the $\downarrow$ symbol are statistically significantly better 
(according to the rank-sum test) compared to the best algorithm in BBOB-2009,
with $p=0.05$ or $p=10^{-k}$ where $k>1$ is the number
following the $\downarrow$ symbol, with Bonferroni correction of 30.

The \IPOPsaACM\ outperforms \IPOPsaACM\ usually by a factor from 2 to 3 on functions with moderate noisy. 
This is the case for Sphere ($f_{101}$,$f_{102}$,$f_{103}$) and Rosenbrock ($f_{104}$,$f_{105}$,$f_{106}$) functions with Gaussian, Uniform and Cauchy noise models.
It seems that the moderate noise only slightly affects the quality of the surrogate model and allows to have a speedup comparable to one of the noise-less case.

On most functions with severe noise the surrogate model usually is not used ($\hat{n}$ oscillates around zero), 
because it gives a very imprecise prediction of the expensive function.
On 20-dimensional $f_{124}$ Schaffer function with Cauchy noise \IPOPsaACM\ requires 
about 7 times more functions evaluations to reach the optimum with $\Delta f_{opt}=10^{-7}$ than IPOP-aCMA-ES (see Fig. \ref{fig:scaling}).
However, according to Table \ref{tab:ERTs20}, the performance for $\Delta f_{opt}=10^{-5}$ is exactly the same for both algorithms, 
therefore, we suppose that the loss of performance for $\Delta f_{opt}=10^{-7}$ can be explained by some influence of surrogate-assisted search on restart conditions. 
We also found that if use default coefficient $c_1$ and $c_{\mu}$ for the covariance matrix update instead of noisy settings of $c_1 /5$ and $c_{\mu}/5$, 
then \IPOPsaACM\ performs as well as IPOP-aCMA-ES. However, the use of default coefficients worsen the results of IPOP-aCMA-ES and \IPOPsaACM\ on other problems. 

We also observe some loss in performance of $f_{125}$ Griewank-Rosenbrock and $f_{128}$ Gallagher functions for $d\leq5$.
For $f_{115}$ we observe the speedup for $d=5$ and loss for $d=20$, probably because of the same reasons as for $f_{124}$.

The \IPOPsaACM\ improves (sometimes insignificantly) the records in dimension 20 on 
$f_{101}$, $f_{102}$, $f_{103}$, $f_{104}$, $f_{105}$, $f_{106}$, $f_{107}$, $f_{109}$, $f_{112}$, $f_{114}$,
$f_{117}$, $f_{118}$, $f_{120}$, $f_{122}$, $f_{123}$, $f_{127}$, $f_{129}$.

\section{CPU TIMING EXPERIMENT}

For the timing experiment the \IPOPsaACM\ was run on noiseless $f_1$, $f_8$, $f_{10}$ and $f_{15}$ functions without self-adaptation of surrogate model hyper-parameters. 
The crucial hyper-parameter for CPU time measurements, the number of training points was set $N_{training} =\left\lfloor 40+4D^{1.7}\right\rfloor$  as a function of dimension $D$ \cite{ACMGECCO2012}.

These experiments have been conducted on a single core with 2.4 GHz under Windows XP using Matlab R2006a.

\begin{figure}[tb]
\begin{center}
  \includegraphics[scale=0.5]{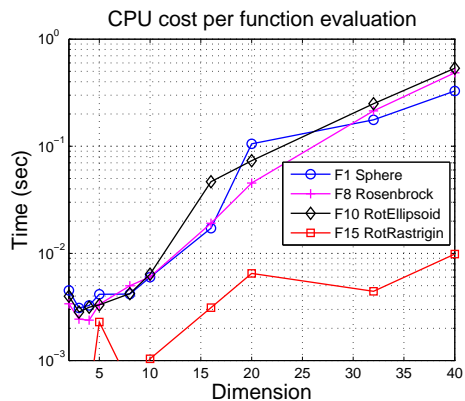}
\end{center}
\caption{\label{fig:time} CPU cost per function evaluation of IPOP-aACM-ES with fixed hyper-parameters.}
\end{figure}

On uni-modal functions the time complexity of surrogate model learning increases cubically in the search space dimension (see Fig. \ref{fig:time}) and quadratically in the number of training points. For small dimensions ($D<10$) the overall time complexity increases super-linearly in the dimension.
The time complexity per function evaluation depends on the population size, because one model is used to estimate the ranking of all points of the population. This leads to a smaller computational complexity on multi-modal functions, e.g. $f_{15}$ Rastrigin function, where the population becomes much larger after several restarts.
The time complexity on noisy functions is more similar to the one of Rastrigin function, because in both cases the large populations are used.

The results presented here does not take into account the model hyper-parameters optimization, where $\lambda_{hyp}$ surrogate models should be build at each iteration, which leads to an increase of CPU time per function evaluation by a factor of $\lambda_{hyp}$. For \IPOPsaACM\ $\lambda_{hyp}$ was set to 20.

\section{Conclusion}

In this paper, we have compared the recently proposed self-adaptive surrogate-assisted \IPOPsaACM\ with the IPOP-aCMA-ES on noisy benchmark problems.
The surrogate-assisted \IPOPsaACM\ algorithm outperforms the original IPOP-aCMA-ES by a factor from 2 to 3 on the functions with moderate noise, 
and usually performs not worse on other functions. 
The \IPOPsaACM\ algorithm improves the records on 15 out of 30 functions in dimension 20.

\section{ACKNOWLEDGMENTS}

The authors would like to acknowledge Anne Auger, Zyed Bouzarkouna, Nikolaus Hansen and Thomas P. Runarsson for their valuable discussions.
This work was partially funded by FUI of System@tic Paris-Region ICT cluster through contract DGT 117 407 {\em Complex Systems Design Lab} (CSDL).

\newcommand{\rot}[2][2.5]{
  \hspace*{-3.5\baselineskip}%
  \begin{rotate}{90}\hspace{#1em}#2
  \end{rotate}}
\newcommand{\includeperfprof}[1]{
  \input{\bbobdatapath #1}%
  \includegraphics[width=0.4135\textwidth,trim=0mm 0mm 34mm 10mm, clip]{#1}%
  \raisebox{.037\textwidth}{\parbox[b][.3\textwidth]{.0868\textwidth}{\begin{scriptsize}
    \perfprofsidepanel 
  \end{scriptsize}}}
}
\begin{table*}\tiny
\mbox{\begin{minipage}[t]{0.48\textwidth}\tiny
\centering
\input{\bbobdatapath pptables_f101_05D} 

\input{\bbobdatapath pptables_f102_05D}

\input{\bbobdatapath pptables_f103_05D}

\input{\bbobdatapath pptables_f104_05D}

\input{\bbobdatapath pptables_f105_05D}

\input{\bbobdatapath pptables_f106_05D}

\input{\bbobdatapath pptables_f107_05D}

\input{\bbobdatapath pptables_f108_05D}

\input{\bbobdatapath pptables_f109_05D}

\input{\bbobdatapath pptables_f110_05D}

\input{\bbobdatapath pptables_f111_05D}

\input{\bbobdatapath pptables_f112_05D}

\input{\bbobdatapath pptables_f113_05D}

\input{\bbobdatapath pptables_f114_05D}

\input{\bbobdatapath pptables_f115_05D}
\end{minipage}

\hspace{3mm}

\begin{minipage}[t]{0.48\textwidth}\tiny
\centering

\input{\bbobdatapath pptables_f116_05D}

\input{\bbobdatapath pptables_f117_05D}

\input{\bbobdatapath pptables_f118_05D}

\input{\bbobdatapath pptables_f119_05D}

\input{\bbobdatapath pptables_f120_05D}

\input{\bbobdatapath pptables_f121_05D}

\input{\bbobdatapath pptables_f122_05D}

\input{\bbobdatapath pptables_f123_05D}

\input{\bbobdatapath pptables_f124_05D}

\input{\bbobdatapath pptables_f125_05D}

\input{\bbobdatapath pptables_f126_05D}

\input{\bbobdatapath pptables_f127_05D}

\input{\bbobdatapath pptables_f128_05D}

\input{\bbobdatapath pptables_f129_05D}

\input{\bbobdatapath pptables_f130_05D}
\end{minipage}}

 \caption{\label{tab:ERTs5}
\bbobpptablesmanylegend{dimension $5$}
 }
\end{table*}

\begin{table*}\tiny
\mbox{\begin{minipage}[t]{0.48\textwidth}\tiny
\centering
\input{\bbobdatapath pptables_f101_20D} 

\input{\bbobdatapath pptables_f102_20D}

\input{\bbobdatapath pptables_f103_20D}

\input{\bbobdatapath pptables_f104_20D}

\input{\bbobdatapath pptables_f105_20D}

\input{\bbobdatapath pptables_f106_20D}

\input{\bbobdatapath pptables_f107_20D}

\input{\bbobdatapath pptables_f108_20D}

\input{\bbobdatapath pptables_f109_20D}

\input{\bbobdatapath pptables_f110_20D}

\input{\bbobdatapath pptables_f111_20D}

\input{\bbobdatapath pptables_f112_20D}

\input{\bbobdatapath pptables_f113_20D}

\input{\bbobdatapath pptables_f114_20D}

\input{\bbobdatapath pptables_f115_20D}
\end{minipage}

\hspace{3mm}

\begin{minipage}[t]{0.48\textwidth}\tiny
\centering

\input{\bbobdatapath pptables_f116_20D}

\input{\bbobdatapath pptables_f117_20D}

\input{\bbobdatapath pptables_f118_20D}

\input{\bbobdatapath pptables_f119_20D}

\input{\bbobdatapath pptables_f120_20D}

\input{\bbobdatapath pptables_f121_20D}

\input{\bbobdatapath pptables_f122_20D}

\input{\bbobdatapath pptables_f123_20D}

\input{\bbobdatapath pptables_f124_20D}

\input{\bbobdatapath pptables_f125_20D}

\input{\bbobdatapath pptables_f126_20D}

\input{\bbobdatapath pptables_f127_20D}

\input{\bbobdatapath pptables_f128_20D}

\input{\bbobdatapath pptables_f129_20D}

\input{\bbobdatapath pptables_f130_20D}
\end{minipage}}
 \caption{\label{tab:ERTs20}
 \bbobpptablesmanylegend{dimension $20$}
}
\end{table*}

%
\bibliographystyle{abbrv}

%
%

\end{document}

%% file: ppdata/bbob_pproc_commands.tex
\providecommand{\bbobpptablesmanylegend}[1]{Expected running time (ERT in number of function evaluations)
                     divided by the respective best ERT measured during BBOB-2009 (given in 
                     the respective first row) for different $\Df$ values in #1. 
                     The central 80\% range divided by two is given in braces. 
                     The median number of conducted function evaluations is additionally given in 
                     \textit{italics}, if $\ERT(10^{-7}) = \infty$.
                     \#succ is the number of trials that reached the final target $\fopt + 10^{-8}$.
                     Best results are printed in bold. }

\providecommand{\bbobpptablesmanylegend}[1]{Expected running time (ERT in number of function evaluations)
                     divided by the respective best ERT measured during BBOB-2009 (given in 
                     the respective first row) for different $\Df$ values in #1. 
                     The central 80\% range divided by two is given in braces. 
                     The median number of conducted function evaluations is additionally given in 
                     \textit{italics}, if $\ERT(10^{-7}) = \infty$.
                     \#succ is the number of trials that reached the final target $\fopt + 10^{-8}$.
                     Best results are printed in bold. }

%% file: BBOB2012_saACMES_noisy.bbl
\begin{thebibliography}{}

\end{thebibliography}


\begin{thebibliography}{10}

\bibitem{augerEvoNum2010}
Z.~Bouzarkouna, A.~Auger, and D.~Ding.
\newblock Investigating the local-meta-model {CMA-ES} for large population
  sizes.
\newblock In {C. Di Chio et al.}, editor, {\em Proc. EvoNUM'10}, pages
  402--411. LNCS 6024, Springer, 2010.

\bibitem{wp200902_2010}
S.~Finck, N.~Hansen, R.~Ros, and A.~Auger.
\newblock Real-parameter black-box optimization benchmarking 2010: Presentation
  of the noisy functions.
\newblock Technical Report 2009/21, Research Center PPE, 2010.

\bibitem{YJin2005}
L.~Graning, Y.~Jin, and B.~Sendhoff.
\newblock Efficient evolutionary optimization using individual-based evolution
  control and neural networks: A comparative study.
\newblock In {\em Proc. ESANN'2005}, pages 27--29, 2005.

\bibitem{hansen2012exp}
N.~Hansen, A.~Auger, S.~Finck, and R.~Ros.
\newblock Real-parameter black-box optimization benchmarking 2012: Experimental
  setup.
\newblock Technical report, INRIA, 2012.

\bibitem{hansen2012noi}
N.~Hansen, S.~Finck, R.~Ros, and A.~Auger.
\newblock Real-parameter black-box optimization benchmarking 2009: Noisy
  functions definitions.
\newblock Technical Report RR-6869, INRIA, 2009.
\newblock Updated February 2010.

\bibitem{HansenECJ01}
N.~Hansen and A.~Ostermeier.
\newblock Completely derandomized self-adaptation in evolution strategies.
\newblock {\em Evolutionary Computation}, 9(2):159--195, 2001.

\bibitem{1830788}
N.~Hansen and R.~Ros.
\newblock Benchmarking a weighted negative covariance matrix update on the
  {BBOB}-2010 noiseless testbed.
\newblock In {\em GECCO '10: Proceedings of the 12th annual conference comp on
  Genetic and evolutionary computation}, pages 1673--1680, New York, NY, USA,
  2010. ACM.

\bibitem{Hoffmann2006IEEE}
F.~Hoffmann and S.~Holemann.
\newblock Controlled model assisted evolution strategy with adaptive
  preselection.
\newblock In {\em International Symposium on Evolving Fuzzy Systems}, pages
  182--187. IEEE, 2006.

\bibitem{Runarsson2011ISDA}
H.~Ingimundardottir and T.~Runarsson.
\newblock Sampling strategies in ordinal regression for surrogate assisted
  evolutionary optimization.
\newblock In {\em Proc. ISDA'2011}, page To appear, 2011.

\bibitem{2006:JastrebskiArnold}
G.~A. Jastrebski and D.~V. Arnold.
\newblock Improving evolution strategies through active covariance matrix
  adaptation.
\newblock In {\em Proc. CEC'2006}, pages 2814--2821, 2006.

\bibitem{kernHansenMetaPPSN06}
S.~Kern, N.~Hansen, and P.~Koumoutsakos.
\newblock Local meta-models for optimization using evolution strategies.
\newblock In {Th. Runarsson et al.}, editor, {\em PPSN IX}, pages 939--948.
  LNCS 4193, Springer, 2006.

\bibitem{KramerInformatica2010}
O.~Kramer.
\newblock Covariance matrix self-adaptation and kernel regression -
  perspectives of evolutionary optimization in kernel machines.
\newblock {\em Fundam. Inf.}, 98:87--106, 2010.

\bibitem{rankSurrogatePPSN10}
I.~Loshchilov, M.~Schoenauer, and M.~Sebag.
\newblock {C}omparison-{B}ased {O}ptimizers {N}eed {C}omparison-{B}ased
  {S}urrogates.
\newblock In J.~K. R.~Schaefer, C.~Cotta and G.~Rudolph, editors, {\em Proc.
  PPSN XI}, pages 364--373. LNCS 6238, Springer, 2010.

\bibitem{ACMGECCO2012}
I.~Loshchilov, M.~Schoenauer, and M.~Sebag.
\newblock {S}elf-{A}daptive {S}urrogate-{A}ssisted {C}ovariance {M}atrix
  {A}daptation {E}volution {S}trategy.
\newblock In {\em GECCO '12: Proceedings of the 14th annual conference on
  Genetic and evolutionary computation}, page to appear, New York, NY, USA,
  2012. ACM.

\bibitem{price1997dev}
K.~Price.
\newblock Differential evolution vs. the functions of the second {ICEO}.
\newblock In {\em Proceedings of the {IEEE} International Congress on
  Evolutionary Computation}, pages 153--157, 1997.

\bibitem{runarssonPPSN06}
T.~P. Runarsson.
\newblock Ordinal regression in evolutionary computation.
\newblock In {Th. Runarsson et al.}, editor, {\em PPSN IX}, pages 1048--1057.
  LNCS 4193, Springer, 2006.

\bibitem{Ulmer2003CEC}
H.~Ulmer, F.~Streichert, and A.~Zell.
\newblock Evolution strategies assisted by gaussian processes with improved
  pre-selection criterion.
\newblock In {\em Proc. CEC'2003}, pages 692--699, 2003.

\end{thebibliography}
